\documentclass{article}

\usepackage{PRIMEarxiv}

\usepackage[utf8]{inputenc} 
\usepackage[T1]{fontenc}    
\usepackage{hyperref}       
\usepackage{url}            
\usepackage{booktabs}       
\usepackage{amsfonts}       
\usepackage{nicefrac}       
\usepackage{microtype}      
\usepackage{lipsum}
\usepackage{fancyhdr}       
\usepackage{graphicx}       
\graphicspath{{media/}}     
\usepackage{xcolor}
\usepackage{hhline}
\usepackage{array}
\usepackage{amsmath}
\usepackage{geometry}
\usepackage{skmath}
\usepackage{comment}
\usepackage{algpseudocode}
\usepackage{algorithm}
\usepackage{xcolor, soul,colortbl}
\sethlcolor{yellow}
\usepackage{placeins}
\usepackage{caption}
\usepackage{float}
\usepackage{comment}
\usepackage[flushleft]{threeparttable}
\usepackage[numbers]{natbib}
\usepackage{multicol}
\usepackage[outdir=./]{epstopdf}

\def\tsc#1{\csdef{#1}{\textsc{\lowercase{#1}}\xspace}}
\tsc{WGM}
\tsc{QE}
\tsc{EP}
\tsc{PMS}
\tsc{BEC}
\tsc{DE}
\title{Balancing Efficiency vs. Effectiveness and Providing Missing Label Robustness in Multi-Label Stream Classification
\thanks{This document is the results of the research
   project funded by the Türkiye Bilimsel ve Teknolojik Araştırma Kurumu (TÜBİTAK). Project No. 122E271. Corresponding author: Fazli Can.}}

\author{
 Sepehr Bakhshi \\
  Bilkent Information Retrieval Group\\
  Department of Computer Engineering\\
  Bilkent University \\
  Ankara, Turkey\\
  \texttt{sepehr.bakhshi@bilkent.edu.tr} \\
   \And
  Fazli Can \\
  Bilkent Information Retrieval Group\\
  Department of Computer Engineering\\
  Bilkent University \\
  Ankara, Turkey\\
  \texttt{canf@cs.bilkent.edu.tr} \\
}

\begin{document}
\maketitle

\begin{abstract}
Available works addressing multi-label classification in a data stream environment focus on proposing accurate models; however, these models often exhibit inefficiency and cannot balance effectiveness and efficiency. In this work, we propose a neural network-based approach that tackles this issue and is suitable for high-dimensional multi-label classification. Our model uses a selective concept drift adaptation mechanism that makes it suitable for a non-stationary environment. Additionally, we adapt our model to an environment with missing labels using a simple yet effective imputation strategy and demonstrate that it outperforms a vast majority of the state-of-the-art supervised models. To achieve our purposes, we introduce a weighted binary relevance-based approach named ML-BELS using the Broad Ensemble Learning System (BELS) as its base classifier. Instead of a chain of stacked classifiers, our model employs independent weighted ensembles, with the weights generated by the predictions of a BELS classifier. We show that using the weighting strategy on datasets with low label cardinality negatively impacts the accuracy of the model; with this in mind, we use the label cardinality as a trigger for applying the weights. We present an extensive assessment of our model using 11 state-of-the-art baselines, five synthetics, and 13 real-world datasets, all with different characteristics. Our results demonstrate that the proposed approach ML-BELS is successful in balancing effectiveness and efficiency, and is robust to missing labels and concept drift.
\end{abstract}

\keywords{Multi-label classification \and Data streams \and Neural networks \and Concept drift \and Missing labels 
}

\section{Introduction}

\label{section:introduction}
In multi-label classification, a model predicts a combination of labels for a single data item. Categorizing a movie into a proper set of genres is a classic example in this domain. Several approaches are proposed for multi-label classification in a static environment, ranging from a kNN classifier to neural networks \cite{zhang2007ml,brinker2007case,crammer2003family,zhang2006multilabel}; however, due to the static nature of these approaches, efficiency is not considered a major factor, and a highly accurate but inefficient model may be an appropriate choice for many tasks in this domain \cite{you2019attentionxml}. In a dynamic environment, data items arrive in temporal order and at a fast rate, necessitating a quick processing paradigm \cite{gama2014survey}. Our experiments in this paper reveal that a significant number of state-of-the-art approaches for data stream multi-label classification models struggle to strike a balance between efficiency and effectiveness. This problem is even more notable when dealing with high-dimensional data. In these datasets, the feature-set size is too large. We consider any dataset with more than 1,000 features to be high-dimensional. 
If a model is not able to keep up with the fast rate of data arrival, it may lose crucial information due to “concept drift”, and high processing costs may result in a decline in the performance of the model in real-world scenarios. Concept drift refers to the changes in the data distribution over time \cite{widmer1996learning}. This change leads to inaccurate predictions \cite{gama2014survey}. For instance, let us suppose a user is interested in two news topics: \{sports, politics\}. After a while, the list of user interests may be updated to include: \{science, art\}. 
If a model fails to keep pace with the fast arrival rate of data, it risks losing vital information due to concept drift, leading to decreased performance in real-world scenarios due to the associated high processing costs. 
\par
Another hurdle in the data stream environment is the lack of labeled data. In a multi-label setting, each data item may have missing labels. It is important to note that the missing label and semi-supervised problem settings are two different problems \cite{liu2021emerging}. In the missing label setting, a label set is partially labeled; however, in a semi-supervised setting \cite{kumar2023online}, a data item either has a set of labels or the whole label set is missing. Despite its importance in real-world scenarios, the missing label problem is unexplored in a non-stationary environment.
\par
\textbf{Motivation.} 
Our motivation is to address three problems in multi-label data stream classification: (1) high processing costs, (2) missing labels, and (3) concept drift. To achieve these goals, we propose a model that can operate efficiently and effectively in both supervised and missing label settings, and handle concept drift. 
\par
Initially, we utilize a lightweight ensemble, which has a low computational burden and reasonable memory usage, in the binary relevance \cite{boutell2004learning} part. As its name implies, in binary relevance, the multi-label problem is transformed into several binary classification problems; however, the label dependencies are not taken into account. To incorporate label dependencies in the binary relevance, we use the predictions of another classifier component as weights for the output of binary relevance predictions. Both ensembles and weighting strategies contribute to the model's ability to quickly adapt to drifts. We also apply a simple yet effective imputation technique that makes our model robust to missing labels. In our approach, we employ the Broad Ensemble Learning System (BELS) \cite{10225305}, a recently proposed model for single-label data stream classification, as our base classifier in both binary relevance and weighting.

\par
\textbf{Contributions.} We introduce ML-BELS (Multi-Label Broad Ensemble Learning System), a novel approach for multi-label data stream classification. To the best of our knowledge, we introduce the first model that handles three challenges simultaneously in a multi-label data stream setting: balancing efficiency and effectiveness, concept drift, and missing labels. We:
\begin{itemize}
\item Balance effectiveness and efficiency by leveraging a lightweight neural network in the binary relevance components;
\item Introduce a novel weighting mechanism based on the predictions of a single-label classifier to consider label dependencies in binary relevance, and incorporate the label cardinality as a signal to trigger the weighting;
\item Ensure robustness to concept drift and missing labels by designing novel strategies to address these problems;
\item Conduct experiments on 18 datasets, compare our results with 11 state-of-the-art baselines and demonstrate that the proposed approach (ML-BELS) is effective and efficient in a stream setting. Our code is publicly available on Github \footnote{\url{https://github.com/sepehrbakhshi/ML-BELS}}. 

\end{itemize}
\textbf{Paper Organization.} In the following, we commence by introducing the problem definition in Section 2, followed by a comprehensive review of related works in Section 3.  Then, we present the proposed approach in Section 4. Next, in the experimental evaluation section (Section 5), we provide a thorough and extensive discussion of the results. Finally, we present our conclusion and possible future directions in Section 6.

\section{Problem Definition}
In a data stream, data items arrive in temporal time order $t$: $\mathcal{D} =\{{d_1}, {d_2}, {d_3},...,{d_t}\}$. Each data item in a multi-label problem has a set of features $\mathcal{X} =\{{x_1}, {x_2}, {x_3},...,x_k\}$ and a set of labels $\mathcal{L} =\{{l_1}, {l_2}, {l_3},...,l_C\}$ \cite{li2015leveraging}.
The model should be able to predict a set of labels $\mathcal{\hat{Y}}$ for each data item $d$. To make a proper prediction for $d_t$, the last $n$ data items are used for learning where $1 \leq n \leq t-1$.
\par
Concept drift refers to changes in the distribution of data over time \cite{lu2018learning}. Based on this definition, in the case of concept drift, we have the following:
\begin{equation}
\nonumber
P_{t-1}(\mathcal{X}_{t-1}, {L}_{t-1})
 \neq P_t (\mathcal{X}_t, {L_t}) 
\end{equation}
There are four main concept drift types: \textit{sudden}, \textit{gradual}, \textit{incremental} and \textit{recurring} \cite{lu2018learning}. As their names imply, in a sudden drift, the distribution of the data changes abruptly and usually results in a severe decline in the performance of the model. In gradual and incremental drift, the new concept replaces the old one gradually and incrementally, respectively. Finally, in recurring drift, an old concept reappears. In this case, the model usually uses the previously trained classifiers and replaces them with the older ones.
\par
In case of a missing label, for each $\mathcal{X}$, the label set size is equal to $(c - u)$, where $u$ stands for the number of missing, and $c$ for the number of available labels. For instance, if we have a total of 100 labels $(c + u=100)$, and the missing label percentage is 90\% (in this example, $u=90$), then the number of available labels $c$ is equal to 10. 
\section{Related Work}
\label{section:related_work}
Approaches that focus on multi-label classification are divided into two main categories: 1) \textit{Algorithm Adaptation (AA)}, and 2) \textit{Problem Transformation (PT)} \cite{buyukccakir2018novel}. In this section, we explore various algorithms developed based on these approaches for a stream environment and analyze their advantages and disadvantages. We also cover studies that address the problem of missing labels in multi-label classification.
\subsection{Algorithm Adaptation (AA)}

In algorithm adaptation, usually, an already existing model for single-label classification is modified so that it can function in a multi-label setting. Numerous models are proposed based on traditional machine learning models like kNN \cite{zhang2007ml, liu2016neighbor, veloso2007multi} and decision trees for this purpose \cite{petrovskiy2006paired, zheng2019survey, pereira2018categorizing}. With the advent of deep learning-based approaches, algorithm adaptation methods are no longer confined to traditional machine learning algorithms. Deep learning-based models focus on more challenging tasks in multi-label classification settings. Extreme multi-label classification \cite{zong2022bgnn} and handling missing labels \cite{chen2022structured} are examples of these tasks.
\par

Tree-based approaches are vastly used in multi-label data stream classification. In \cite{read2012scalable}, a variation of the Hoeffding Tree is used for multi-label classification, where multi-label classifiers serve as the leaves of the tree. Another approach based on tree structure is proposed in \cite{osojnik2017multi}, where a combination of problem transformation and algorithm adaptation techniques is utilized. Specifically, a multi-label classification method is transformed into a multi-target regression task, and an incremental Structured Output Prediction Tree (iSOUP-Tree) is used for multi-target regression.
\par
Lazy learning strategies are extensively employed in multi-label settings. Inspired by the idea of self-adjusting memory (SAMkNN) \cite{losing2016knn} in single-label stream classification, the authors of \cite{roseberry2018multi} and \cite{roseberry2019multi} propose kNN approaches with self-adjusting memory for multi-label classification. Focusing on handling concept drift in data streams using self-tuning the parameters, the authors propose MLSAkNN \cite{roseberry2021self}. Adaptive Ensemble of Self-Adjusting Nearest Neighbor Subspaces (AESAKNNS) \cite{alberghini2022adaptive} ensures the diversity of classifiers by training them on separate subsets of features and training instances. Online Dual Memory (ODM) \cite{wang2021multi} uses short-term memory to adapt to concept drift. Long-term memory is used additionally to handle recurring concepts. A recent kNN-based named Aging and Rejuvenating kNN
(ARkNN) \cite{roseberry2023aging} focuses on efficiency and achieves competitive results compared to that of its predecessors; however, its efficiency decreases drastically when examined on high-dimensional data.
\par

Apart from kNN and tree-based approaches, in \cite{nguyen2019multi}, authors propose a Bayesian approach that considers label correlation, and label and feature correlation into account. 
Inspired by their outstanding performance, ensemble methods are also considered for multi-label classification in a non-stationary environment. GOOWE \cite{bonab2018goowe} (Geometrically Optimum Online Weighted Ensemble) is an ensemble approach designed for single-label classification tasks. GOOWE-ML \cite{buyukccakir2018novel} applies GOOWE in a multi-label setting.

\subsection{Problem Transformation (PT)}
Unlike the algorithm adaptation technique that applies the modification to the algorithm, in problem transformation, the modification is applied to the data by transforming it into multiple single-label classification tasks \cite{zhang2013review}. There are two widely studied techniques for problem transformation: 1) Binary Relevance (BR) \cite{boutell2004learning} and 2) Label Powerset (LP) \cite{tsoumakas2010random}. 
\par
In binary relevance, for each label, a classifier or a set of classifiers (in the case of an ensemble approach \cite{wei2009mining,qu2009mining}) is assigned, and these classifiers decide whether a data item belongs to a class or not.
Binary relevance has a major drawback that leads to incorrect class predictions: it fails to consider the correlation between class labels, treating each label as independent of the others. Classifier Chains \cite{read2011classifier} addresses this issue by creating a stacked set of classifiers, where the prediction of each classifier is added to the feature set and fed to the next classifier until all class labels are covered. One challenge posed by this technique is determining the optimal order in which to employ the classifiers in the stacked algorithm.

In contrast to the binary relevance method that requires at least one classifier for each class label, a single classifier is sufficient for the label powerset approach. This method considers each combination of class labels as a single class label, transforming the problem into a single-label classification task with $2^{\mathcal{L}}$ labels; however, a major drawback of this approach is that many class labels have very few or no instances, resulting in an imbalanced dataset that negatively impacts the model's performance, especially for large label sets. Compared to the binary relevance approach, the label powerset is more efficient.
In \cite{junior2017label,junior2019pruned}, two label powerset-based approaches are proposed for a non-stationary stream environment. Recently, a similar method to label powerset has been proposed, which transforms the label set into an integer and uses regression to determine the final set \cite{gulcan2022binary}.
\subsection{Handling Missing Labels}
Handling missing labels is a less explored domain in the multi-label setting. Most of the available models that address the scarcity of labels are designed for semi-supervised settings, where each data instance either receives or misses the whole set of labels; however, In the missing labels settings, a subset of each data item is labeled. As indicated in \cite{wu2014multi}, we consider the set of labels for each data instance as $y \in \{-1,0,+1\}$, where $-1$, $+1$, and $0$ mean the label at index $c$ is negative, positive, or not available, respectively. In \cite{liu2021emerging}, the authors utilize this definition and categorize multi-label learning with missing labels into three categories:

\begin{itemize}
\item Explicit Missing Labels: missing labels are explicitly annotated as missing,
\item Implicit Missing Labels: missing labels are implicit and in the form of false positives and negatives,
\item Partial Missing Labels: potential true positives are marked, but the labels may contain false positives.
\end{itemize}

In this paper, we focus on multi-label learning with \textit{Explicit Missing Labels}.
In a static environment, the missing labels problem is a well-studied topic \cite{he2019joint,rastogi2021multi}. In \cite{wu2018multi}, the authors consider mixed dependencies utilizing a graph strategy. Graph-based methods are among the dominant strategies used for handling missing labels \cite{yang2016improving, yu2017unified,huang2023multi}.
\par
Embedding is another strategy that is widely used in the missing labels scenario. For instance, in \cite{wang2014binary}, the authors code both the features and the labels in binary and propose an optimization approach to learn the binary codes. 
\par
An alternative method is \textit{label imputation}. This strategy involves imputing (assigning) a label to the missing ones. Using this technique, the authors in \cite{ma2019label} recover the missing labels in two steps using imputation.

To the best of our knowledge, there is no model capable of handling both the missing labels and concept drift problems in a stream environment in the same setting as ours. There are a few works that consider semi-supervised multi-label classification in a stream environment \cite{chu2019co,qiu2022semi,xu2017dynamic,li2021online}; however, their proposed approaches are out of the scope of this work.

\section{ML-BELS: Proposed Approach}
Broad Learning System (BLS) \cite{chen2017broad} is an exceptionally efficient neural network model when compared to deep models. First and foremost, the model uses the least square solution which is considerably efficient compared to gradient descent \cite{chen2017broad}. Additionally, BLS does not have a deep structure; instead, the model utilizes only three connected layers. Initially, in the feature mapping layer, an improved feature representation is acquired by using random weights and a sparse autoencoder. Next, the output of the feature mapping layer is fed to the enhancement layer where a $tanh$ function is utilized. Eventually, the outputs of these two layers are concatenated to solve the least square problem and generate the appropriate weights for predicting the upcoming data. BLS is designed for a non-stationary environment, is not able to handle concept drift, and is not suitable for data streams. 
\par 
To use BLS in a data stream environment, Broad Ensemble Learning System (BELS) \cite{10225305} uses two techniques. First, it uses a more comprehensive feature mapping layer by updating it at each timestep. This culminates in generating a feature representation that encompasses the whole data including the latest timestep. Next, an ensemble approach is designed to handle the concept drift. 

A single feature mapping and enhancement layer is used for all the data,
and the ensemble consists of output layer instances. Each output layer instance consists of several output nodes. With this technique, the model operates efficiently since the feature mapping and enhancement layers need only one update at each timestep, and having an ensemble of these layers is unnecessary. In the ensemble, each instance of the output layer is added or removed based on its accuracy performance. Each removed instance is kept in a pool (later illustrated in Figure \ref{fig:concept_drift}) and can return to the learning process if its accuracy passes a threshold. Using this technique, the model is able to handle concept drift in a single-label classification scenario.
\par
\begin{figure*}[t]
    \centering
    \includegraphics[width=0.65\textwidth,keepaspectratio]{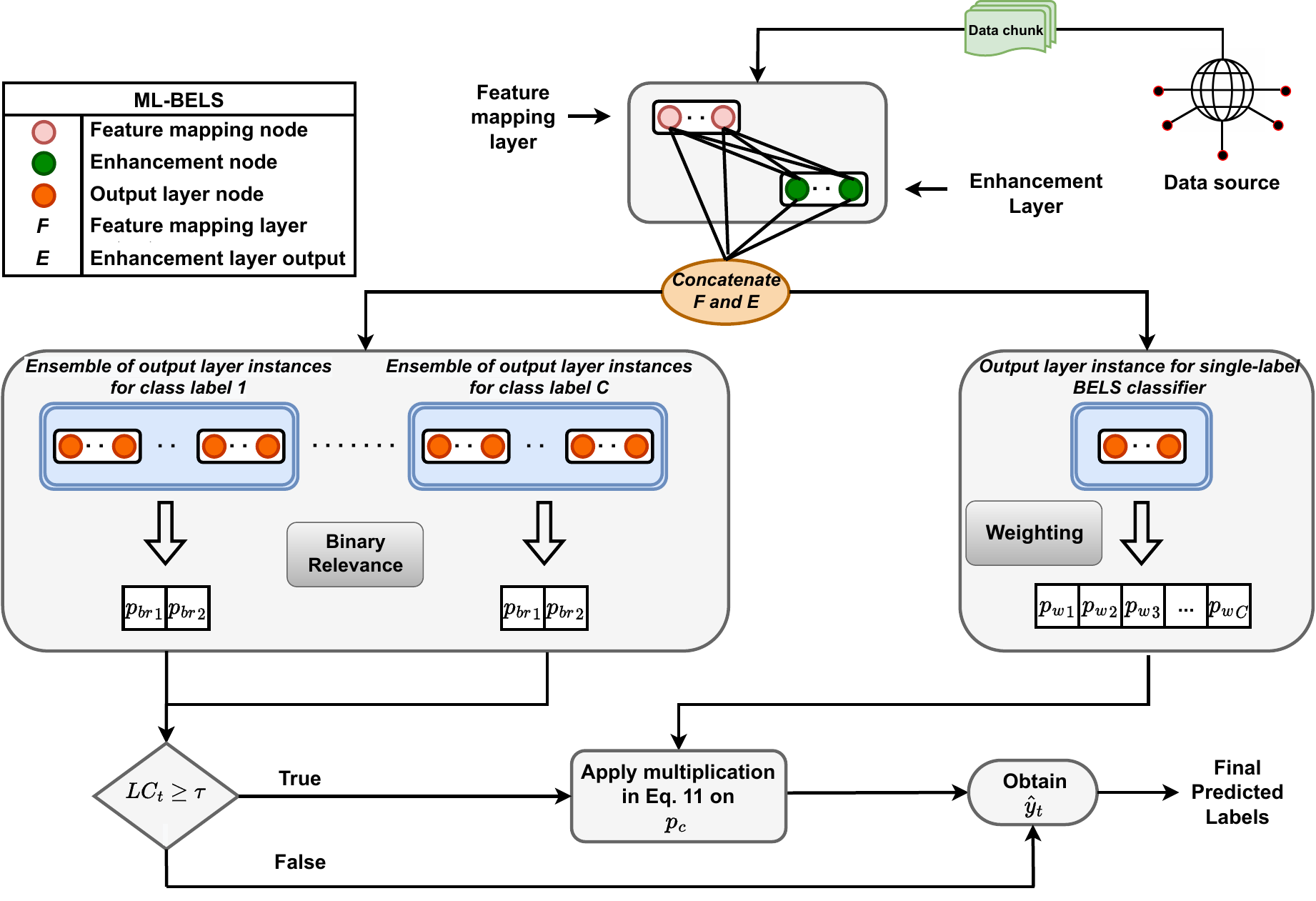}
    \caption{Schematic representation of binary relevance and weighting mechanism used in ML-BELS. At each timestep, a chunk of data is used to generate the new features. Then it is passed to the binary relevance and weighting components. The output is the final predicted labels.}
    \label{fig:MLBELS_BR}
\end{figure*}

 
In this section, our model is presented in four distinct parts: binary relevance, label dependency incorporation, concept drift adaptation, and handling missing labels. Notations that are used in the formulas are given in Table \ref{tbl:notations}. The overall schema of our model is shown in Figure. \ref{fig:MLBELS_BR}. Its components are explained in the following sections.

\par\begin{table*}[t]
    \centering  
    \caption{ML-BELS notation table }
    \begin{tabular}{c l| c l}
        \hline
        Symbol & Meaning & Symbol & Meaning  \\
        \hline 
        $n$& Number of data items in a chunk &  $W_w$ & Single-label BELS output weights \\
        $C$ & Number of class labels &  $p_{br}$ &Predictions for each label in BR phase\\
        $e$ & Ensemble size &$m$ & Concatenation of F and E of test phase\\
        $F$ &Feature mapping layer output of the training phase & $p_c$& Concatenation of BR predictions\\
        $E$ & Enhancement layer output of the training phase & $p_w$ & Single-label BELS prediction\\
        $M$ & Concatenation of F and E & $\hat{y}_t$  & Final predicted labels at timestep $t$\\
        $BR$ & Binary Relevance
        & $LC_t$ & Label cardinality at timestep $t$\\
        $W_F$ & Random weights of feature mapping layer & $\tau$ & Label cardinality threshold\\
        $W_E$ & Random weights of enhancement layer & $\theta$& Default accuracy threshold in the ensemble \\
     ${W_{br}}$ & Output weights of BELS components in BR  & $Y_t$& Labels at timestep $t$ \\
        \hline
    \end{tabular}
    \label{tbl:notations}
\end{table*}
\subsection{Decreasing Computational Cost of Binary Relevance}
Binary relevance adds a high computational burden to the model. This problem is even more observable when the feature and label sizes increase. Choosing a fast approach is essential in such cases since it alleviates the inefficiency of binary relevance. 
\par
For this purpose, we choose BELS as the base classifier \cite{10225305}. In BELS, the ensemble is built using the output layer nodes of BELS as its components, where only a least square solution is needed to predict the labels. We use this feature of BELS and build our binary relevance using only the output layer nodes. This means that at each timestep, the feature mapping and enhancement layer is calculated once based on the input data. Then the outputs of these layers are fed into the binary relevance components. We have an ensemble of output layer instances for each class label, and each instance consists of several output nodes. \par
Let us assume that $F$ and $E$ are the outputs of feature mapping and enhancement layers, respectively. $F$ and $E$ are matrices with proper dimensions based on the number of their nodes \cite{10225305,chen2017broad}.
For a dataset with $C$ class labels and an ensemble of $e$ output layer instances, our ensemble has a total of ($C \times e$) output layer instances. 
\par
For generating the feature mapping layer $F$, a linear transformation function is applied on the feature set using a set of random features $W_{F}$ and a bias $\beta_{F}$ \cite{10225305,chen2017broad}:
\begin{equation}
    {F} = \phi_i(X_t W_{F} +\beta_{F})
    \label{eq_pre1}
\end{equation}
 Then an activation function (tanh in our approach) is applied on the feature mapping nodes \cite{10225305,chen2017broad}:
\begin{equation}
    {E} = tanh(F W_{E} +\beta_{E})
    \label{eq_pre2}
\end{equation}
where $W_{E}$ and $\beta_{E}$ are random weights and bias values generated with proper dimensions, respectively. 
  After generating $F$ and $E$, their output is concatenated. ${M}$ is the result of this concatenation.
\begin{equation}
    {M} = [F, E]
    \label{eq1}
\end{equation}
${M}$ is a matrix with $(n \times (d_E + d_F))$ dimensions where $d_F$ and $d_E$ are the dimensions of $F$ and $E$, respectively, and $n$ is the chunk size. Next, the result of this concatenation is fed into the ensemble. Each label has a separate ensemble with \textit{e} output layer node sets.
 In order to solve the least square for each component of binary relevance, we need to find $W_{i}^j$ ($ i \in \{1,2,3,...,C\}$, and $ i \in \{1,2,3,...,e\}$) for each output layer instance in that component. Each label is also represented in one hot encoding before updating the weights where $Y_i \in [0,1]$. 

 Then the least square is solved as follows \cite{10225305,chen2017broad}:
\begin{equation}
    {W_{{br}_i}^j} = {(\lambda I + {M}^T{M}}^{-1}) {{M}^T}Y_{i}
   \label{eq2}
\end{equation}
In this equation, $I$ is an identity matrix, and $\lambda$ is the regularization parameter \cite{chen2017broad}. ${W_{{br}_i}}^j$ is a matrix with $((d_E + d_F) \times 2)$ dimensions. Since we use binary relevance, each label vector length is equal to two (one for representing the occurrence of the label in the actual label set and the other for the opposite case).
\par
After finding a proper ${W_{{br}_i}}$ in the training phase, trained feature mapping and enhancement layers are used to generate a suitable set of features for the incoming test data. We name the concatenation of the test results of the feature mapping and enhancement layer as $m$.

To generate the set of predictions, each member of the ensemble uses ${W_{{br}_i}}$ and $m$ as follows:
\begin{equation}
    p_{{br}_i}= m {W_{{br}_i}^j}
   \label{eq3}
\end{equation}
The prediction results of each member of the ensemble are then summed up to form a single output for each ensemble with size $(n \times 2)$. Finally, we concatenate all the predictions to make a three-dimensional matrix with dimensions equal to $(C \times n \times 2 )$. 
\begin{equation}
     \forall i \in \{1,..., C\}, 
     p_{c}= [p_{1}, ..., p_{i}]
   \label{eq4}
\end{equation}
\par

\subsection{Incorporating Label Dependencies by Predictions}
One of the major drawbacks of the binary relevance approach is its inability to incorporate label dependencies. To solve this issue, we generate a set of weights for each component of binary relevance. The weights are the predictions of a single-label classifier. With this strategy, the prediction considers both label-label and feature-label dependencies.
\par

After calculating the predictions of each ensemble in the binary relevance part and concatenating them, we use the $M$ matrix ($M$ is calculated in the binary relevance step) and feed it to the output layers (Eq. \ref{eq5}).
This output layer acts like a single-label classifier, and its output is a set of prediction values for each class.
\begin{equation}
    W_w = {(\lambda I + {M}^T{M})}^{-1} {M}^T Y
   \label{eq5}
\end{equation}
We also use the same $m$ of the binary relevance part. The following formula returns the predictions of the single-label BELS which is a matrix of size $(n \times C)$:
\begin{equation}
    p_w = m W_w
   \label{eq6}
\end{equation}

Subsequently, we apply \textit{min-max normalization} on the rows of p:

\begin{equation}
p_w = \frac{p_w - min(p_w)}{max(p_w) - min(p_w)}
\label{minmax_1}
\end{equation}
\par
We use \textit{Label Cardinality} (LC) as the signal to apply the predictions as weights. 
In a multi-label problem setting, label cardinality is defined as the average number of labels for each data instance and formulated as follows \cite{tsoumakas2007multi}:
\begin{equation}
    LC(D) = \frac{1}{ C} \sum_{i=1}^{ C}  { Y_i,}
    \label{eq9}
\end{equation}
where $C$ is the number of class labels.
\par
It is trivial that label dependencies become more complex when the label cardinality is higher.
  Based on this fact, we use the label cardinality as a decision rule and decide whether the model should apply weights on the output of binary relevance or not. To do so, we update the label cardinality of the model during the learning process. If the label cardinality of the dataset at each chunk becomes more than a predefined threshold $(\tau)$, the signal for using the weights is triggered. 
  \par

\newlength{\commentlen}
\algnewcommand\algorithmicinput{\textbf{Input:}}
\algnewcommand\algorithmicoutput{\textbf{Output:}}
\algnewcommand\Input{\item[\algorithmicinput]}%
\algnewcommand\Output{\item[\algorithmicoutput]}%
\begin{algorithm}[t]
\caption{ML-BELS: BR and Weight Assignment}

\setlength{\commentlen}{10ex}
\begin{algorithmic}[1]
\Require  {$D$: Data stream}
\Ensure {${\hat{y}}_t$: prediction of the model at step $t$}
\While{$D$ has more instances}
\State{$X_t$ = Data chunk at timestep $t$}
\State{$Y_t$ = Labels at timestep $t$}
\Procedure{Test}{$X_t$, $W_{br}$, $W_w$, $LC_t$, $\tau$}
  
    \State{Calculate $F$ and $E$ using $X_t$}
    \State{$m =[F, E] $}
    \For{each class i }
        \State{Calculate $p_{{br}_i}$} \Comment{\makebox[\commentlen][l]{Eq. \eqref{eq3}}}
    \EndFor
    \State{Calculate $p_w$} \Comment{\makebox[\commentlen][l]{Eq. \eqref{eq6}}}
     
    \If{$LC_t \geq \tau$}
        
        \State{$p_c$= Apply weights on BR outputs} \Comment{\makebox[\commentlen][l]{Eq. \eqref{eq7}}}
        
    \EndIf

    \State{$P_c$=[$p_1$,..., $p_C$]} 
    \State{$\hat{y}_t$ = Final decision \hfill \Comment{\makebox[\commentlen][l]{Eq. \eqref{eq8}}}}
    
    \State{Return $\hat{y}_t$}
\EndProcedure
\Procedure{Train}{$X_t$, $Y_t$}
    \State{Update $F$ and $E$ using $X_t$} \Comment{\makebox[\commentlen][l]{Eq.\eqref{eq_pre1},\eqref{eq_pre2}}}
    \State{$M =[F, E] $} 
    \For{each class $i$, and output layer instance $j$}
        \State{Calculate $W_{{br}_i}^j$} \Comment{\makebox[\commentlen][l]{Eq. \eqref{eq2}}}
    \EndFor
    \State{Calculate $W_w$} \Comment{\makebox[\commentlen][l]{Eq. \eqref{eq5}}}
   \State{Calculate label cardinality $LC_{t}$} \Comment{\makebox[\commentlen][l]{Eq. \eqref{eq9}}}
    \State{Return $W_{br}$, $W_w$, $LC_t$ }
\EndProcedure
\EndWhile
\end{algorithmic}
\label{alg:1}
\end{algorithm}
\par
 
  Eq. \ref{eq7} is used for applying the weights. In this equation, the normalized predictions of the single-label BELS classifier are used as weights for the output of binary relevance. The score of each class label is multiplied by its respective ensemble output in the binary relevance part.

\newcommand{\Hquad}{\hspace{0.4em}} 
\begin{equation}
\begin{aligned}
    \forall i \in \{1,..., C\}, \forall j \in \{1,..., n\}, \forall k \in \{0,1\}, \Hquad
    p_{c}=
    \begin{cases}
        (p_{c_{i,j,k}}) (1-p_{i,j}), \Hquad \Hquad   k = 0 \\
        (p_{c_{i,j,k}}) (p_{i,j}),\Hquad \Hquad \Hquad \Hquad \quad   k = 1 
    \end{cases}
\end{aligned}
\label{eq7}
\end{equation}
In the last step, we find our final predictions using a simple comparison:
\begin{equation}
\begin{aligned}
    \forall i \in \{1,..., C\}, \forall j \in \{1,..., n\}, \Hquad  
    \hat{y}_{j,i}=
    \begin{cases}
        0, \Hquad \Hquad   (p_{c_{i,j,0}}) > (p_{c_{i,j,0}}) \\
        1,\Hquad \Hquad   (p_{c_{i,j,0}}) \leq (p_{c_{i,j,0}})\\
    \end{cases}
\end{aligned}
\label{eq8}
\end{equation}

\par The procedure of the ML-BELS model including the test and train phases introduced in Sections 4.1 and 4.2 is shown in Algorithm \ref{alg:1}.

\subsection{Selective Concept Drift Adaptation}
For handling concept drift, we use a combination of ensemble learning and weighting. 
\par

\textbf{Ensemble-based approach.}
 Available ensemble methods for concept drift handling rely on the whole set of labels for reacting to drift. In our approach, we react to drifts inside each binary relevance component. With this approach, our model is capable of handling class-level drifts and only the component responsible for that class label is updated for the drift. We call it a \textit{Selective Concept Drift Adaptation} mechanism. 
A similar approach is used in MLSAkNN \mbox{\cite{roseberry2021self}}. In this work, the authors do not use an ensemble, and the parameters of the kNN are adjusted with regard to concept drift; however, our approach utilizes a lightweight ensemble for each class label and adjusts to drifts according to the performance of the ensemble on each class label separately.

\par
We apply this technique utilizing the concept drift handling mechanism of BELS. In BELS, an ensemble of output layer instances is used to effectively handle concept drift. The model adds or removes these output layer instances based on their accuracy. In our approach, for each class in the label set, we have an ensemble of output layer instances. We add or remove the instances based on their example-based accuracy. If the accuracy of these output layer instances is lower than a threshold ($\theta$), then they are moved to a pool of removed output layer instances. Next, a new one, or an output layer instance from the pool is brought back to the learning process. 
\par
\begin{figure*}[t]
    \centering
    \includegraphics[scale = 0.35]{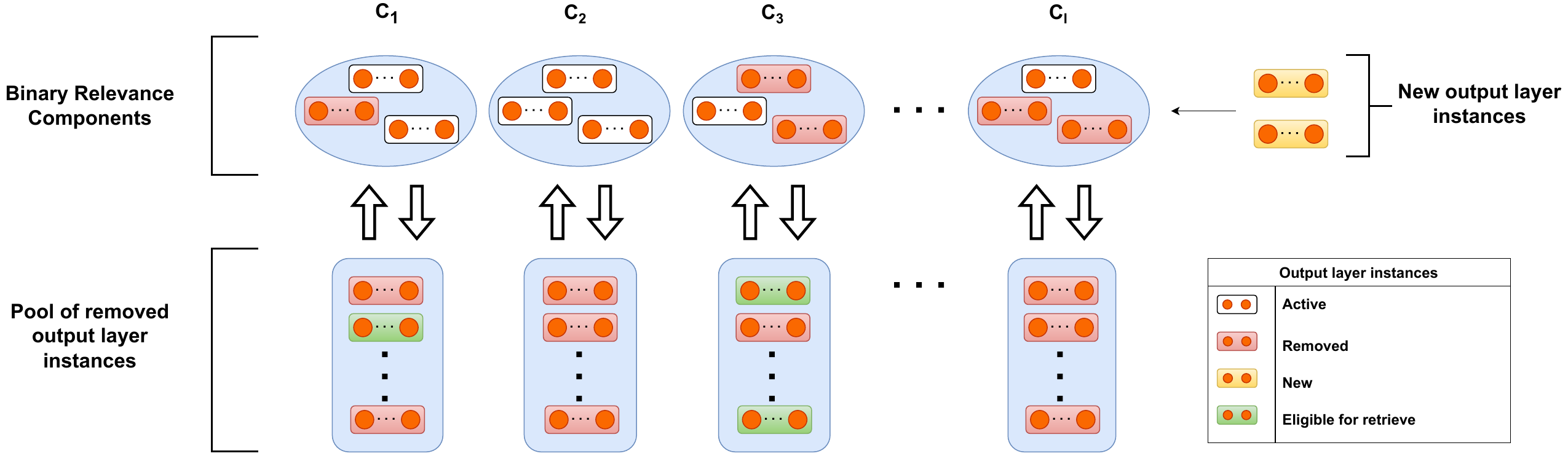}
    \caption{Concept drift handling in ML-BELS using an ensemble of output layer instances. Each label in binary relevance has an ensemble of output layer instances. Each instance is a set of output layer nodes. The output layer instances are removed or replaced with a new or recovered old output layer instance based on its example-based accuracy. }
    \label{fig:concept_drift}
\end{figure*}

The detailed steps of the concept drift handling mechanism of BELS are demonstrated in Figure \ref{fig:concept_drift}. Each ensemble of output layer instances in Figure \ref{fig:MLBELS_BR}, handles drift for each class label.
Each output layer instance in each ensemble may be in one of the four states:
\begin{itemize}
    \item \textbf{Active}: It is participating in the learning process, and is used as one of the decision-makers in the majority voting.
    \item \textbf{Removed}: It is removed from the learning process and moved to the pool of removed output layer instances as of the next chunk.
    \item \textbf{New}: A new output layer instance replaced by one of the removed instances in the binary relevance component.
    \item \textbf{Eligible for retrieval}: The accuracy of the previously removed output layer instance passes the threshold, is eligible for retrieval and is replaced with one of the removed instances in the binary relevance component.
\end{itemize}
\par

For instance, for the first class label, there is one output layer instance with an accuracy lower than the threshold, and it is ready to be replaced with either a new or a previously removed one from the pool. The priority is with the ones in the pool; however, if the classifiers in the pool do not pass the threshold check (like the ones in $C_l$), we add new output layer instances to the ensemble and move the removed ones to the pool. The pool size is fixed, and if it passes a certain size, we remove the older ones from the pool permanently. The procedure of ensemble-based concept drift handling is shown in Algorithm \ref{alg:2}.
\par
\textbf{Weighting.}
We introduced our weighting mechanism to incorporate label dependencies in Section 4.2. As mentioned earlier, the weighting system basically encodes the feature-feature and label-label dependencies and applies them as weights on the output of binary relevance. The weighting is dynamically updated and reflects the changes in case of drift, resulting in a faster concept drift adaptation in datasets with high label cardinality. The weights are the second part of our selective concept drift handling while dealing with datasets with high label cardinality. The effect of the proposed weighting mechanism on datasets with drift is studied later in an ablation analysis in Section 5.4.

\begin{algorithm}[H]
\caption{ML-BELS: Concept Drift Adaptation}

\begin{algorithmic}[1]

\Require  {BR components, $\theta$}
\Ensure {BR components after drift adaptation}
\For{i= 0 to $i\leq C$}
    \For{j=0 to $j \leq e$}
        \If{Acc. of output layer instance $j$ in $BR_i$ $\leq$ $\theta$} 
        \State{Remove the instance $j$ from $BR_i$}
        \State{Move $j$ to pool}
        \If{Pool is full}
            \State{Remove the first instance from pool}
        \EndIf
        \State{Check pool for eligible output layer instances}
        \If{Pool has eligible instance}
            \State{Replace it with the removed one}
        \Else
            \State{Add new instance to the $BR_{i}$}
        \EndIf
    \EndIf    
    \EndFor
\EndFor
\end{algorithmic}
\label{alg:2}
\end{algorithm}

\par

\begin{figure*}[h]
    \centering
    \includegraphics[scale = 0.5]{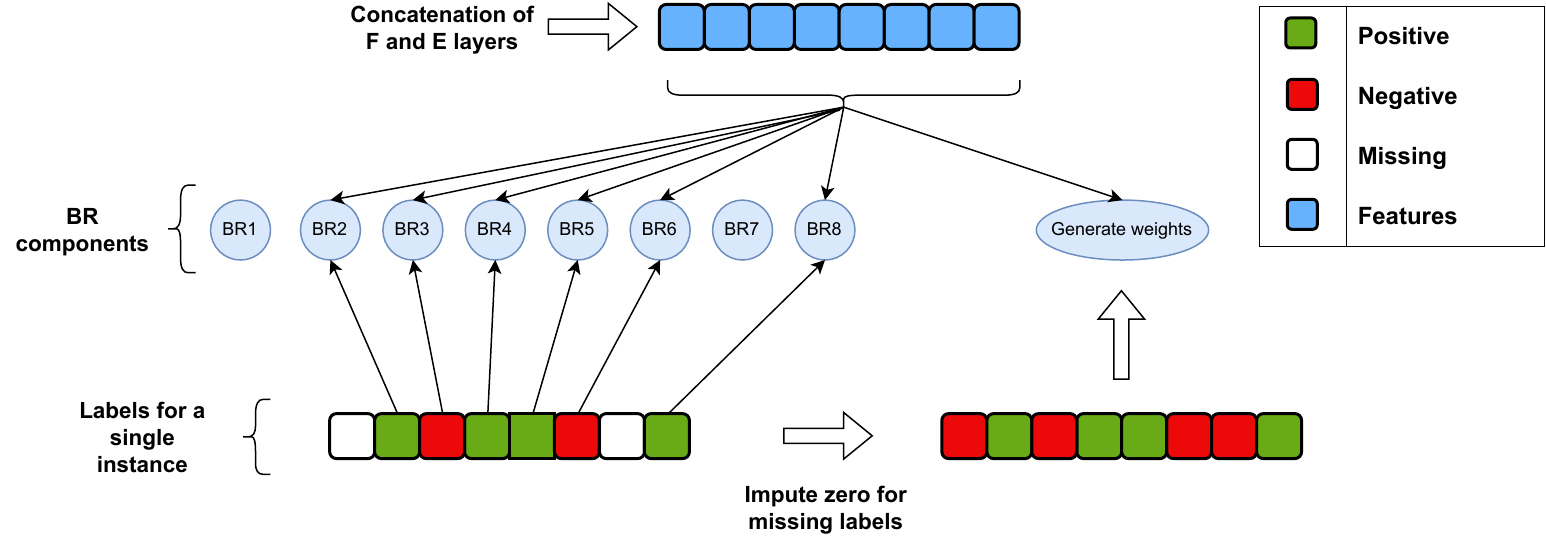}
    \caption{Missing label handling in ML-BELS. The above example data instance contains two missing, four positive, and two negative labels. Classifiers in BR are not considered when the label is missing (in the figure, there is no connection between the label and the corresponding BR component). The labels provided for single-classifier BELS are replaced with negative labels.}
    \label{fig:missing_label_handling}
\end{figure*}
\subsection{Handling Missing Labels by Label Imputation}
To have a robust classifier in the presence of explicit missing labels, we adapt our model to this setting so that it can function effectively in such a scenario.  \par
The binary relevance components are exclusively fed with labeled instances. For the classifier that generates weights for the binary relevance components, we provide it with the complete set of labels; however, this time we consider the missing labels as negative. The intuition behind this imputation choice is that the majority of multi-label datasets have sparse labels in real-world scenarios \cite{sun2019partial}. Considering unassigned labels as negative has been studied in the literature and is a naive yet effective approach for handling missing labels \cite{bucak2011multi,sun2010multi}. Since we do not have access to all the labels in this setting, the approximate label cardinality based on the available labels is used as a trigger for weighting. 
\par
Figure \ref{fig:missing_label_handling} demonstrates the modification we make in our model for handling the missing labels. The figure illustrates an example instance with eight labels. The instance 
 has two missing labels, two negative labels, and four positive labels. The binary relevance components related to the class labels with the missing labels are not trained with that instance. This approach offers three distinct advantages: \begin{itemize}
     \item Since the missing labels are removed from the training process, we have total confidence in the remaining labeled examples, and only the number of training instances for each binary relevance component is decreased in this way.
     \item Considering most of the unassigned labels are negative labels \cite{sun2019partial}, this removal acts like an undersampling that balances the positive and negative classes.
     \item By imputing negative labels and feeding them to the model to update the weights, we capture the feature-label and label-label dependencies for available positive labels. 
 \end{itemize}

 \par

\begin{algorithm}[h]
\caption{ML-BELS: Missing Label Handling}

\begin{algorithmic}[1]

\Require  {BR components, Single-label BELS}
\Ensure {Trained BR components and single-label BELS}
\For{i= 0 to $i\leq C$}
    \If{Label is available} 
        \State{Train $BR_i$ on class label i}
    \Else
      \State{Impute negative label instead of the missing one}
    \EndIf
\EndFor
\State{Train Single-label BELS on the imputed label set}
\end{algorithmic}
\label{alg:3}
\end{algorithm}

By applying these strategies to address missing labels, our approach not only achieves robustness but also delivers results that are competitive with the baselines in their fully supervised mode. The procedure for training the model with missing labels is presented in Algorithm \ref{alg:3}.

\section{Experimental Evaluation}
To demonstrate the effectiveness and efficiency of our approach, and compare our model with the state-of-the-art
baselines in terms of different criteria, we design an experimental evaluation that supports our claims. In the following, we first take a glance
at our experimental setup and introduce the datasets used in our experiments.
Then, we compare our model with the baselines, conduct an ablation study,
and analyze the effect of hyperparameters on the performance of ML-BELS.
\subsection{Experimental Environment}
\subsubsection{Baselines}
In our experiments, we use \textit{interleaved-test-then-train} approach \cite{gama2009issues}. In this technique the incoming data or data chunk is first used for testing and then for training.

\par
We compare our model with 11 state-of-the-art baselines that cover various techniques discussed in Section \ref{section:related_work}. 
Among the kNN-based approaches, we choose the two most recent works AESAKNNS\footnote{https://github.com/canoalberto/AESAKNNS} \cite{alberghini2022adaptive} and ARkNN \footnote{https://github.com/canoalberto/ARkNN} \cite{roseberry2023aging}. From the GOOWE-based approaches proposed in \cite{buyukccakir2018novel}, we include the fastest (GOPS) and the most accurate (GOCC) variations of GOOWE-ML as our baselines. We also use classic baselines for multi-label data stream classification including EaBR \cite{read2012scalable}, EaCC \cite{read2012scalable}, EaPS \cite{read2012scalable},  EBR \cite{read2011classifier}, ECC \cite{read2011classifier}, EPS \cite{read2008multi} and EBRT \cite{osojnik2017multi}. Three of these models (EaBR, EaCC, EaPS) use ADWIN \cite{bifet2007learning} concept drift detector. The baselines are implemented using WEKA \cite{witten2002data}, MOA \cite{bifet2010moa}, and MEKA \cite{read2016meka} libraries and we use the publicly available implementations of these models \cite{buyukccakir2018novel}.\footnote{https://github.com/abuyukcakir/gooweml}
\par

\subsubsection{Datasets}
The information regarding our datasets is provided in Table \ref{tbl:datasets}. We use 15 real-world datasets with different feature and label sizes, low and high label cardinality, and different densities. Using synthetic datasets is considered in recent works \cite{gulcan2023unsupervised,wang2021multi}. To show the effectiveness of our approach in the presence of drift, we generate five synthetic datasets including three drift types (sudden, gradual, and recurring), and noise. In Table \ref{tbl:datasets} (U), (A), (G), and (R) stand for Unknown, Abrupt, Gradual, and Recurring drift types, respectively. There are two drift points in each dataset. The first number in synthetic datasets' name shows the number of class labels and the second number indicates the noise percentage. The MOA library is utilized for generating synthetic datasets. The drifts are induced by modifying the label cardinality and label dependency \cite{wang2021multi}. The SEA generator is used as the binary generator. 
\par

\renewcommand{\arraystretch}{1}
\setlength{\tabcolsep}{4.3pt}
\begin{table*}[t]
\centering\footnotesize
\caption{Summary of datasets}
\begin{tabular}{l r r r l l l r r}

\hline
Dataset &   Features  &    Labels &      Instances &   Type & Domain & Drift type& LC & LD   \\
\hline



CAL500 & 68& 174& 502 & Real & Music&U & 26.044& 0.150\\

CHD49 & 49& 6& 555 & Real & Medicine&U & 2.580& 0.430\\


Enron   & 1,001 & 53 &  1,702 & Real  & Text & U & 3.378 & 0.064\\


IMDB    & 1,001 & 28 &  120,919 & Real &Text &
U &2.000 &0.071 \\


Medical    & 1,449 & 45 &  978 & Real &Text &
U &1.245 &0.028\\

Ohsumed    & 1,002 & 23 &  13,930 & Real &Text &
U &1.663 &0.072\\

Reuters    & 500 & 103 &  6,000 & Real &Text &
U &1.462 &0.014\\

Slashdot    & 1,079 & 22 &  3,782 & Real &Text &
U &1.181 &	0.054\\

Stackex-chess    & 585 & 227 &   1,675 & Real &Text &
U &2.411 &	0.011\\ 

TMC    & 500 & 22 &  28,600 & Real &Text &
U &2.220 &	0.101\\ 


Yahoo-Computers & 34,096 & 33 & 12,444& Real &Text &
U &	1.507 &	0.046\\

Yahoo-Society & 31,802 & 27 & 14,512& Real &Text &
U &	1.670 &	0.062\\

Yeast    & 103 & 14 &  2,417& Real &Biology &
U &	4.237 &	0.303\\

A-10-20    & 3 & 10 &  20,000& Syn &Numeric &
A &	3.293 &	0.329\\

A-20-20   & 3 & 20 &  20,000& Syn &Numeric &
A &	6.442 &	0.322\\

G-10-10   & 3 & 10 &  20,000& Syn &Numeric &
G &	3.293 &0.329\\

G-20-10   & 3 & 20 &  20,000& Syn &Numeric &
  G&	6.442 &	0.322\\

A-R-15-10    & 3 & 20 &  20,000& Syn &Numeric &
A \& R&4.352 &	0.290\\ 
\hline
\end{tabular}
\label{tbl:datasets}
\end{table*}
\begin{table*}[t]
\centering\footnotesize
\caption{List of hyperparameters and their default value}
\begin{threeparttable}
\begin{tabular}{|l|l|r|}

\hline
Hyperparameter &    Definition &  Default Value  \\
\hline

e & No. of output layer instances
for each ensemble in the model & 3\\
\hline
$d_F$ & No. of feature mapping nodes & 25\\
\hline
$d_E$ & No. of enhancement nodes & 1\\
\hline
-\tnote{*} & Pool size (previous instances of output layer) & 100 \\
\hline
$\theta$ & Default accuracy threshold in the ensemble & 0.5\\
\hline
$\tau$ & Default threshold
of label cardinality for triggering the weighting & 1.5\\
\hline

\end{tabular}
\begin{tablenotes}
    \item[*] No symbol is used for this in the text.
    \end{tablenotes}
\end{threeparttable}
\label{tbl:param}
\end{table*}

\subsubsection{Setting}
We use a default set of parameters given in their respective papers in all of our experiments. The chunk size for each dataset is chosen from a predefined set of values, namely $[50, 100, 250, 500, 1000]$, based on the size of the dataset. A similar approach is also used in \cite{buyukccakir2018novel}. 

\par
 As for the hyperparameters of our base classifier BELS, we use 3 output layer instances for each ensemble in our model $(e = 3)$. In the feature mapping layer and enhancement layer, 25 and one node are used, respectively. The example-based accuracy threshold ($\theta$) for removing an output layer instance in the ensembles of the binary relevance part is set to (0.5). We set the pool size to 100 for each BR component. For ML-BELS, we minimize the number of hyperparameters and only add one more hyperparameter ($\tau$) to the above-mentioned ones. The default threshold of label cardinality for triggering the weighting is set to ($\tau = 1.5$).
The hyperparameters, a short description, and their default values are provided in Table \ref{tbl:param}. 
\par
For the missing label scenario, we randomly remove labels from each chunk of data. In our experiments, we keep two different portions of labeled data: {30\% and 10\%}. For each dataset, we run the experiments five times and report the average value for each evaluation metric.
\par
Our computer is equipped with an Intel(R) Xeon(R) Gold 5118 CPU
@ 2.30 GHz, 128GB RAM, and the operating system is Ubuntu 18.04.4 LTS.

\subsubsection{Effectiveness Measures}
We report the results for the most common effectiveness metrics in evaluating multi-label data stream classification including example-based accuracy, F1 score, and micro-average F1 \cite{zhang2013review,gulcan2023unsupervised}. Two other metrics for multi-label classification evaluation are subset accuracy and Hamming score. Since subset accuracy is extremely strict and the reported accuracy results are low or in some cases zero, we do not report the results for this metric. In the GOOWE-ML paper \cite{buyukccakir2018novel}, the authors demonstrate that the Hamming score is not a suitable choice for measuring multi-label classifiers' performance in stream settings; therefore, we also do not include the Hamming score in our report. Finally, we show that our model is both efficient and effective for multi-label stream classification by reporting the runtime of the model and comparing it with the baselines. 

\par

\renewcommand{\arraystretch}{0.99}

\begin{table*}
    \centering
    \caption{ Example-based accuracy, example-based F1-score, Micro-f1 score, and runtime results. Runtime is reported in seconds for processing 10 data instances. Results marked with $\dagger$ are not obtained due to memory overflow, thus the average value for each metric and their corresponding ranks are calculated based on the available results. The best results for each row are in bold}
    \resizebox{1\textwidth}{!}{
    \begin{tabular}{l l r r r r r r r r r r r r r r}
        \hline 
        Datasets & & ML-BELS&  ML-BELS(30\%)&  ML-BELS(10\%) & AESAKNNS& ARkNN &GOCC& GOPS&  EBR & EaBR  &ECC & EaCC & EPS &EaPS &  EBRT \\
        
        \hline
        
         
         


        CAL500 &Acc.& 0.238 &0.204&  0.182 & \textbf{0.272}& 0.226&0.270&0.244&0.212& 0.211&0.212 & 0.212& 0.134& 0.019&0.000\\
        
         & F1& 0.398 &0.343&  0.316 & \textbf{0.428} & 0.368&\textbf{0.428}&0.401&0.349& 0.349&0.349& 0.349&0.229& 0.035&0.000\\
         
         &Micro F1&  0.379 &0.335&  0.307 & \textbf{0.423} & 0.364&0.420&0.392&0.343& 0.342&0.343& 0.343&0.254&0.045&0.000\\
         
        &Runtime& 0.210 &0.170& 0.150 & 2.385 & \textbf{0.068}&4.280&0.272&4.178& 6.236&5.087& 8.120&0.456&0.332&3.721\\
        \hline

        CHD49 &Acc.& 0.498 &0.432&  0.390 & \textbf{0.568 }& 0.443& 0.500&0.473&0.496&0.493&0.493& 0.493& 0.462&0.463&0.000\\
        
         & F1& 0.642 &0.564&  0.547 & \textbf{0.693} & 0.584&0.642&0.615&0.638&0.634&0.638&0.638&0.608&0.609&0.000\\
         
         &Micro F1&  0.643 &0.584&  0.540 & \textbf{0.702} & 0.605&0.645&0.620&0.643&0.640&0.639&0.639&0.611&0.615&0.000\\
         
        &Runtime&0.023  &\textbf{0.019}&  \textbf{0.019} & 0.123& 0.025&0.050 &0.034&0.050&0.068&0.059&0.073&0.030&0.056&0.022\\
        \hline

        
         
         
        
        Enron &Acc.&0.371&0.337&0.303& 0.334&\textbf{0.373}&0.351&0.294&0.342&0.308&0.298&0.289&0.302&0.260&0.059\\
        
        & F1&\textbf{0.511}&0.460&0.429&0.471&0.501&0.494&0.448&0.468&0.425&0.414&0.408&0.397&0.347&0.061\\
        
        &Micro F1& \textbf{0.508}&0.480&0.431&0.453&0.479&0.488&0.417&0.440&0.411&0.403&0.395&0.358&0.330&0.037\\
        
        &Runtime& 0.060&0.047&0.042&1.538&\textbf{0.009}& 7.994&0.407&7.004&7.639&6.964&8.643&0.146&0.484&1.084\\
        
        \hline
        IMDB &Acc.&\textbf{0.249}&0.248&0.246& 0.189&0.163&0.138&0.148&0.055& 0.024&0.012& 0.001&0.011& 0.019&0.000\\
        
        &F1&\textbf{0.361}&0.357&0.348&0.288&0.238&0.221&0.256&0.075& 0.031&0.016& 0.001&0.015& 0.026&0.000\\
        
        &Micro F1& \textbf{0.351}&0.349&0.345&0.270&0.219&0.228&0.252&0.099& 0.041&0.025& 0.001&0.018& 0.033&0.000\\
        
        &Runtime&0.029&0.020&0.017&0.590&\textbf{0.004}&4.230&0.989&4.289& 2.677&4.198& 2.835&0.165& 0.605&0.526\\
        
        \hline
        
        
        
        
        
        Medical &Acc.&\textbf{ 0.584} &0.454& 0.311 & 0.418 & 0.461&0.444&0.266 &0.386&0.384&0.386&0.386&0.338&0.314&0.002\\
        
         & F1& \textbf{0.613} &0.452&  0.335 & 0.496 & 0.520&0.482&0.356 &0.416&0.414&0.414&0.414&0.370&0.344&0.002\\
         
         &Micro F1&  \textbf{0.618 }&0.501&  0.366 & 0.515 & 0.518&0.593&0.312&0.548&0.546&0.549&0.549&0.373&0.354&0.003\\
         
        &Runtime& 0.075 &0.061&  \textbf{0.054} & 1.957& 0.348&9.650&0.326&9.265&11.778&9.295&14.733&0.198&0.391&0.896
        \\
        \hline
        Ohsumed & Acc.&\textbf{0.267}&0.261&0.241&0.009&0.056  & 0.267& 0.186& 0.191 & 0.169 &0.180 &0.004 &0.135 &0.113 &0.049 \\
        
        &F1&\textbf{0.369}&0.352&0.308&0.017&0.116&0.345&0.308&0.230&0.202&0.217&0.004&0.160&0.134&0.056\\
        
        &Micro F1&0\textbf{.349}&0.339&0.310&0.020&0.119&0.401&0.280&0.294&0.266&0.280&0.007&0.197&0.171&0.076\\
        
        &Runtime&0.022&0.018&\textbf{0.017}&1.681&0.086 &4.230 & 0.650 & 3.413 & 3.431 & 2.246 & 1.632 & 0.225& 1.020 & 0.854\\
        
        \hline
        
        Reuters &Acc.&\textbf{0.377}&0.342&0.294&0.274&0.343&0.134&0.167&0.098&0.056&0.093&0.004&0.131&0.158&0.000\\
        
        &F1&\textbf{0.411}&0.343&0.279&0.338&0.398&0.160&0.266&0.106&0.059&0.098&0.004&0.136&0.164&0.000\\
        
        &Micro F1&0.373&0.347&0.305&\textbf{0.379}&0.347&0.205&0.214&0.141&0.081&0.134&0.007&0.153&0.184&0.000\\
        
        & Runtime&0.126&0.090&0.081&3.509&\textbf{0.003}&7.021&0.260&5.205&5.080&5.443&5.045&0.090&0.328&1.581\\
        
        \hline
        
        Slashdot &Acc.&\textbf{0.361}&0.308&0.260&0.154&0.318  &0.047 &0.157  &0.020 &0.016 &0.018 &0.018 &0.070 &0.044 &0.001\\
        
        & F1&\textbf{0.371}&0.296&0.252&0.238&0.369 & 0.050 & 0.255 &  0.023 & 0.018 & 0.020 & 0.020 & 0.075 & 0.047& 0.001\\
        
        &Micro F1& \textbf{0.362}&0.314&0.271&0.228&0.357&0.084&0.225&0.041& 0.033& 0.037&0.037 & 0.106&0.074&0.000\\
        
        &Runtime&0.030&0.025&\textbf{0.023}&1.001&0.206  & 3.545 & 0.710 & 3.105 & 3.742 & 3.052 & 4.105 & 0.587 & 0.726 & 0.610\\

        \hline
        Stackex-chess &Acc.& \textbf{0.199}&0.166&  0.129 & 0.049 & 0.098&0.008&0.107&0.007&0.006&0.003&0.003&0.080&0.036&0.000 \\
        
         & F1& \textbf{0.309} &0.245&  0.186 & 0.087 & 0.161 &0.011&0.201&0.011&0.011&0.006&0.006&0.112&0.051&0.000\\
         
         &Micro F1&  \textbf{0.268}&0.231&  0.183 & 0.089 & 0.154&0.018&0.160&0.014&0.014&0.008&0.008&0.125&0.067&0.000  \\
         
        &Runtime&0.222  &0.169& \textbf{0.147}  & 6.638 & 0.187& 27.919&0.362&55.376&44.989&25.578&66.419&0.227&0.390&2.725\\
        \hline
        
        TMC & Acc.&0.494&0.490&0.482&0.434& 0.469
        & 0.510 & 0.478  &0.520 & \textbf{0.529} &0.511 &0.516 &0.469 &0.481 &0.007\\
        
        &F1&0.652&0.643&0.622&0.577&0.599& \textbf{0.664}& 0.633& 0.654& 0.661& 0.643& 0.646& 0.590& 0.598& 0.007\\
        
        &Micro F1&0.609& 0.605&0.598&0.550&0.563&0.629&0.594&0.638&\textbf{0.640}&0.631&0.632&0.566&0.577&0.008\\
        
        &Runtime&0.022&0.017&\textbf{0.015}&1.397&0.024  & 1.714 & 0.089 & 1.580& 1.089 & 0.948 & 2.024 & 0.050 & 0.133 & 0.147\\

        \hline

        Yahoo-Computers &Acc.& \textbf{0.444 }&0.441&  0.417 & 0.290 & 0.398&$\dagger$&0.161&$\dagger$&$\dagger$&$\dagger$&$\dagger$&0.172&0.352&$\dagger$   \\
        
         & F1& \textbf{0.515} &0.506&  0.461 & 0.376 & 0.463&$\dagger$&0.277&$\dagger$&$\dagger$&$\dagger$&$\dagger$&0.221&0.402&$\dagger$\\
         
         &Micro F1& \textbf{ 0.454} &0.450&  0.430 & 0.336& 0.407&$\dagger$&0.253&$\dagger$&$\dagger$&$\dagger$&$\dagger$&0.282&0.402&$\dagger$\\
         
        &Runtime& 0.063 &0.049&  \textbf{0.046} & 1.954 & 4.483&$\dagger$&43.356&$\dagger$&$\dagger$&$\dagger$&$\dagger$&9.796&20.160&$\dagger$\\
        \hline

        Yahoo-Society &Acc.& \textbf{0.402} &0.396&  0.378& 0.249 & 0.296&$\dagger$&0.184&$\dagger$&$\dagger$&$\dagger$&$\dagger$&0.162&0.150 &$\dagger$  \\
        
         & F1& \textbf{0.474 }&0.465&  0.432 & 0.352 & 0.371&$\dagger$&0.315&$\dagger$& $\dagger$&$\dagger$&$\dagger$&0.189&0.169&$\dagger$\\
         
         &Micro F1&  \textbf{0.412} &0.407&  0.391 & 0.305 & 0.312&$\dagger$&0.268&$\dagger$&$\dagger$&$\dagger$&$\dagger$&0.205&0.196&$\dagger$\\
         
        &Runtime& 0.050&0.044&  \textbf{0.042} & 1.980 & 3.137&$\dagger$ &17.758&$\dagger$&$\dagger$&$\dagger$&$\dagger$
        &15.992&48.321&$\dagger$\\

         \hline

        Yeast&Acc.&0.503&0.460&0.404&0.481&0.454& 0.509& \textbf{0.519}  &0.502 &0.502 &0.493 &0.495 &0.480 &0.474 &0.000  \\
        
        &F1&0.632&0.578&0.522&0.650&0.584  & 0.652& \textbf{0.655}& 0.638& 0.638 &0.632&0.633& 0.601& 0.596& 0.000\\
        
        &Micro F1& 0.635&0.601&0.552&0.622&0.581&\textbf{0.640}&\textbf{0.640}&0.631&0.632&0.625&0.627&0.604&0.600&0.000\\
        
        &Runtime&0.024&0.020&\textbf{0.018}&0.233&0.021& 0.175& 0.045 & 0.128 & 0.185 & 0.213 & 0.196 & 0.041 & 0.056 & 0.413\\
        \hline
         
        A-10-20&Acc.&\textbf{0.419}&0.409&0.397&0.401&0.286&0.379&0.404&0.252&0.251&0.316 & 0.332 &0.358&0.308&0.000\\
        
        &F1&\textbf{0.597}&0.581&0.559&0.569&0.419&0.527&0.560&0.366&0.367&0.449 & 0.469 &0.499&0.440&0.000\\
        
        &Micro F1& \textbf{0.577}&0.567&0.554&0.553&0.418&0.529&0.559&0.407&0.403&0.457
        & 0.476 & 0.504&0.452&0.000\\
        
        & Runtime&0.011&0.008&0.007&0.098&\textbf{0.001}&0.008&0.002&0.006&0.009&0.008&0.009&\textbf{0.001}&0.003&0.013\\
        
        \hline

        A-20-20& Acc.&\textbf{0.442}&0.436&0.412&0.373&0.309&0.435&\textbf{0.442}&0.389&0.384&0.433&0.428&0.421&0.421&0.000\\
        
        &F1& \textbf{0.605}&0.594&0.559&0.544&0.457&0.583&0.592&0.542&0.538&0.581&0.577&0.569&0.570&0.000\\
        
        &Micro F1& 0.598&0.593&0.571&0.528&0.463&0.592&\textbf{0.599}&0.550&0.546&0.590&0.586&0.578&0.579&0.000\\
        
        & Runtime&0.020&0.015&0.013&0.265&\textbf{0.001}&0.023&0.012&0.012&0.018&0.021&0.028&0.002&0.018&0.026\\
        
        \hline
        
        G-10-10&Acc.&\textbf{0.419}&0.411&0.395&0.400&0.288&0.381&0.407&0.254&0.251&0.334&0.347&0.358&0.347&0.000 \\
        
        &F1&\textbf{0.596}&0.584&0.558&0.568&0.420&0.525&0.563&0.364&0.361&0.467&0.483&0.498&0.482&0.000\\
        
        &Micro F1&\textbf{0.577}& 0.569&0.551&0.551&0.420&0.527&0.561&0.413&0.409&0.475&0.489&0.505&0.489&0.000\\
        
        &  Runtime&0.011&0.008&0.007&0.097&\textbf{0.001}&0.010&0.002&0.006&0.007&0.008&0.010&\textbf{0.001}&0.003&0.014\\

        \hline

        G-20-10&Acc.&\textbf{0.443}&0.433&0.412&0.369&0.311&0.436&0.440 &0.388&0.387&0.433&0.429&0.407&0.413&0.000\\
        
        &F1&\textbf{0.606}&0.592&0.558&0.540&0.459&0.584&0.590&0.541&0.539&0.582&0.577&0.555&0.561&0.000\\
        
        &Micro F1&\textbf{0.599}& 0.591&0.571&0.524&0.465&0.593&0.596&0.550&0.549&0.591&0.586&0.565&0.570&0.000\\
        
        &  Runtime&0.019&0.015&0.012&0.271&\textbf{0.001}&0.023&0.012&0.013&0.020&0.021&0.026&0.002&0.007&0.026\\
        
         \hline

        A-R-15-10&Acc.&\textbf{0.405}&0.398&0.379&0.364&0.270&0.319&0.403 &0.288&0.294&0.255&0.262&0.392&0.367&0.000\\
        
        &F1&0.543&0.530&0.498&0.535&0.402&0.456&\textbf{0.548}&0.421&0.426&0.377&0.387&0.526&0.502&0.000\\
        
        &Micro F1&\textbf{0.557}& 0.550&0.532&0.514&0.412&0.471&0.553&0.450&0.457&0.400&0.430&0.541&0.518&0.000\\
        
        &  Runtime&0.016&0.011&0.011&0.176&\textbf{0.001}&0.015&0.003&0.010&0.014&0.014&0.015&0.002&0.004&0.020\\
        \hline

        Avg. &Acc.&\textbf{0.395}&0.368&0.335&0.312&0.309&0.321&0.304&0.275&0.267&0.279&0.264&0.271&0.263&0.007\\
        &F1&\textbf{0.511}&0.471&0.432&0.432&0.413&0.427&0.436&0.365&0.355&0.369&0.351&0.353&0.338&0.008 \\
        &Micro F1& \textbf{0.493}&0.467&0.434&0.420&0.400&0.441&0.416&0.388&0.376&0.387&0.363&0.364&0.348&0.008 \\
        &Runtime&0.057&0.045&\textbf{0.040}&1.439&0.478&4.430&3.627&5.853&5.436&3.947&7.120&1.557&4.058&0.792\\
       \hline
        
        Avg. Rank&Acc.&\textbf{1.722}&4.389&6.722&6.611&7.444&4.875&5.444&7.688&9.063&8.188&8.813&8.444&8.889&13.875\\
        &F1&\textbf{1.778}&4.389&7.167&5.556&7.389&4.750&4.889&7.938&9.000&8.375&8.875&8.833&9.111&13.875\\
        &Micro F1&\textbf{1.833}&4.556&6.500&6.500&8.167&4.500&5.222&7.750&8.938&8.000&8.563&8.667&9.167&13.875\\
       &Runtime& 5.556&3.500&2.444&10.333&\textbf{2.667}&11.188&5.556&9.063&10.563&10.250&12.063&4.167&6.611&9.250\\
    \hline
        
    \end{tabular}
    }
    \label{tbl:accuracy}
\end{table*}

\begin{figure*}
    \centering
    \begin{tabular}{c c c}
\includegraphics[width=0.3\columnwidth]{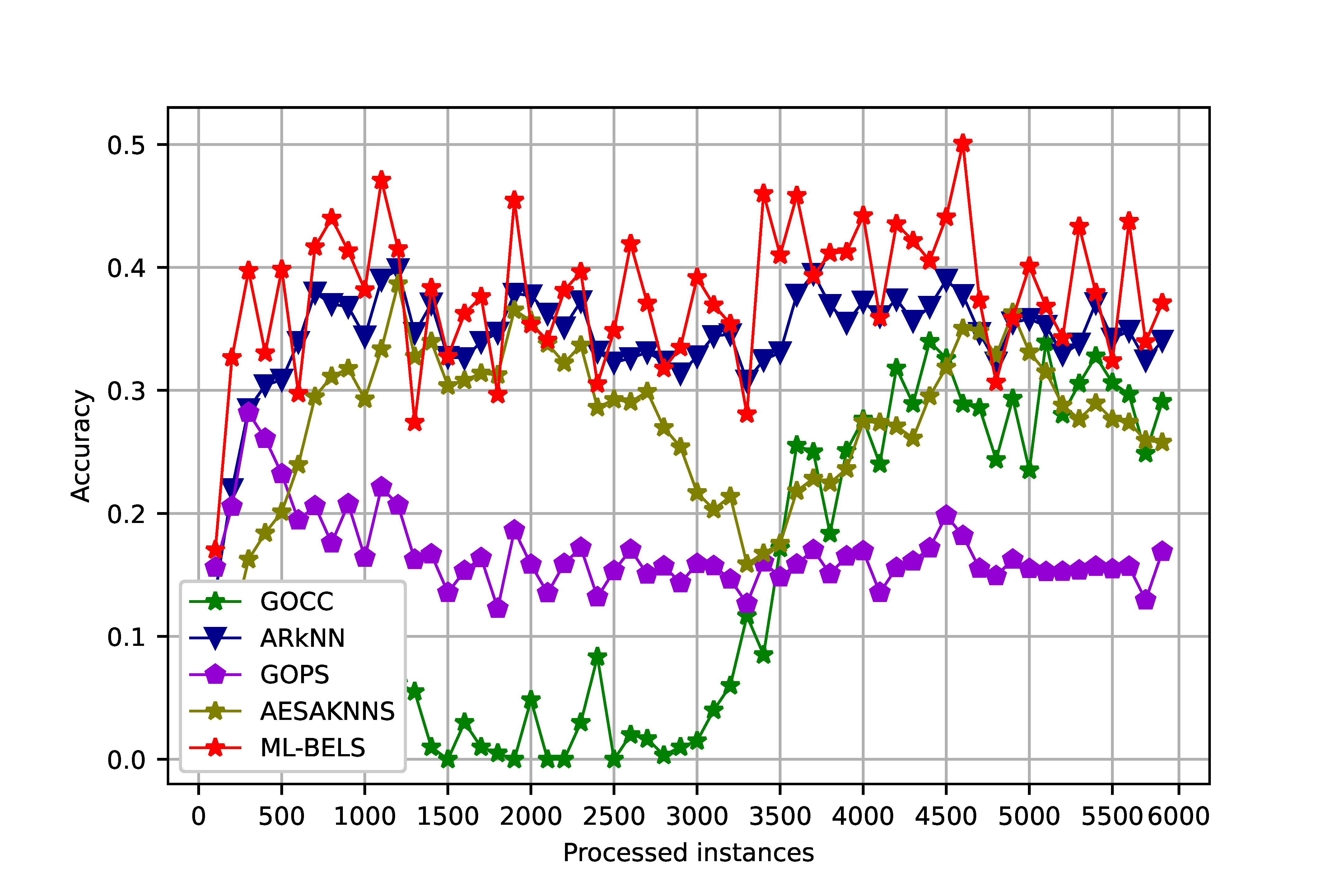}
    & 
\includegraphics[width=0.3\columnwidth]{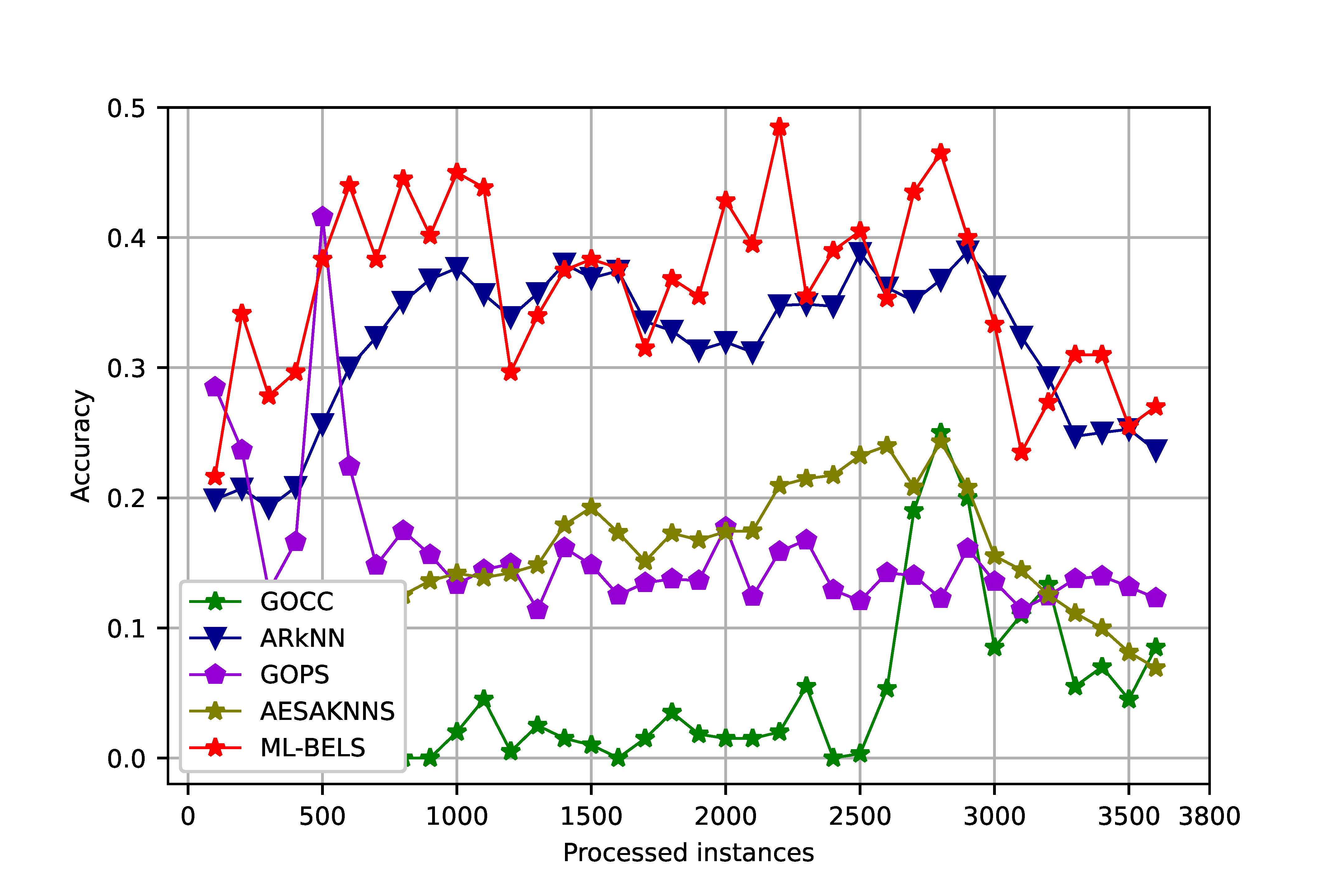}&  \includegraphics[width=0.3\columnwidth]{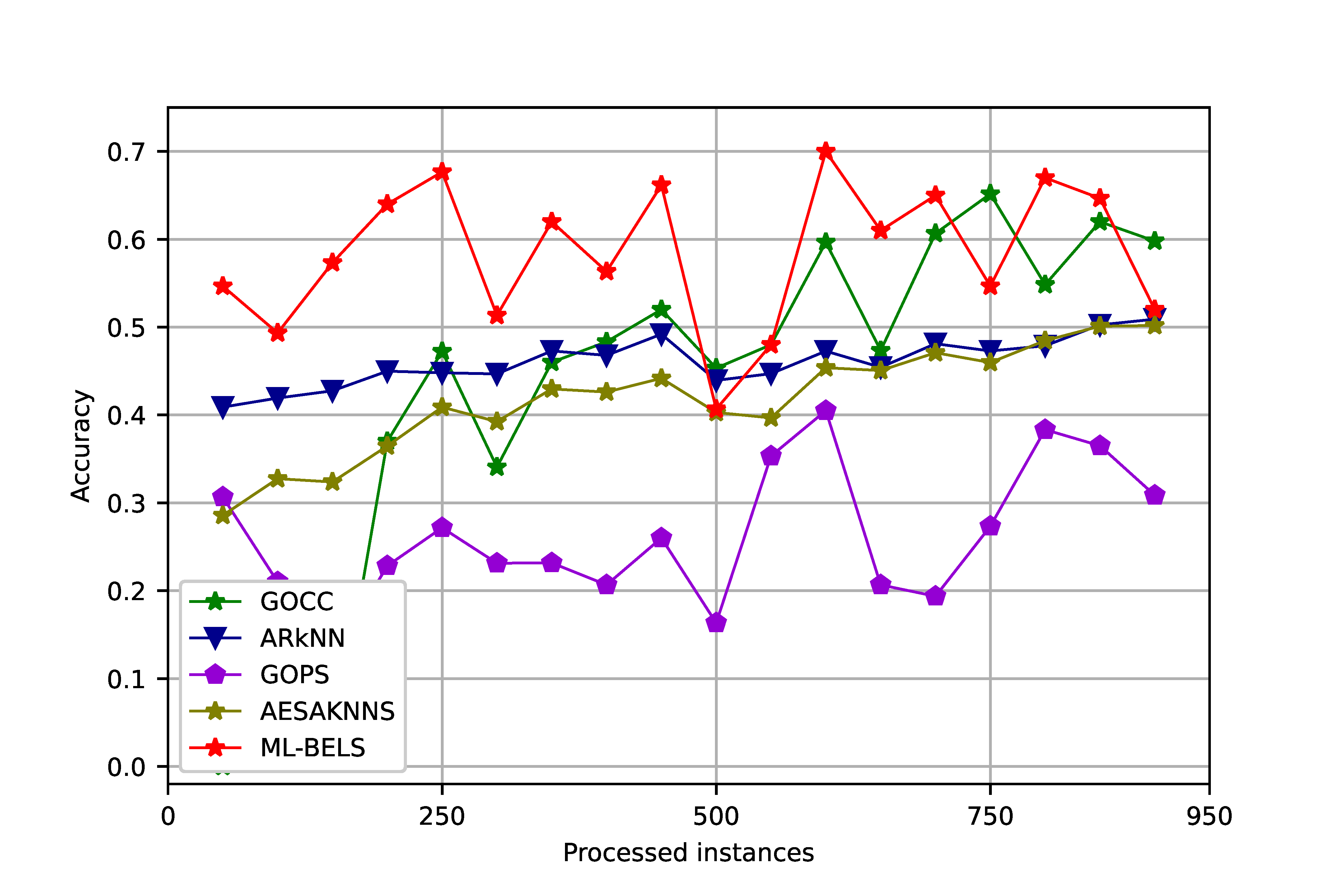}
    \\
    (a) Reuters &
    (b) Slashdot&
    (c) Medical 
    \\
\includegraphics[width=0.3\columnwidth]{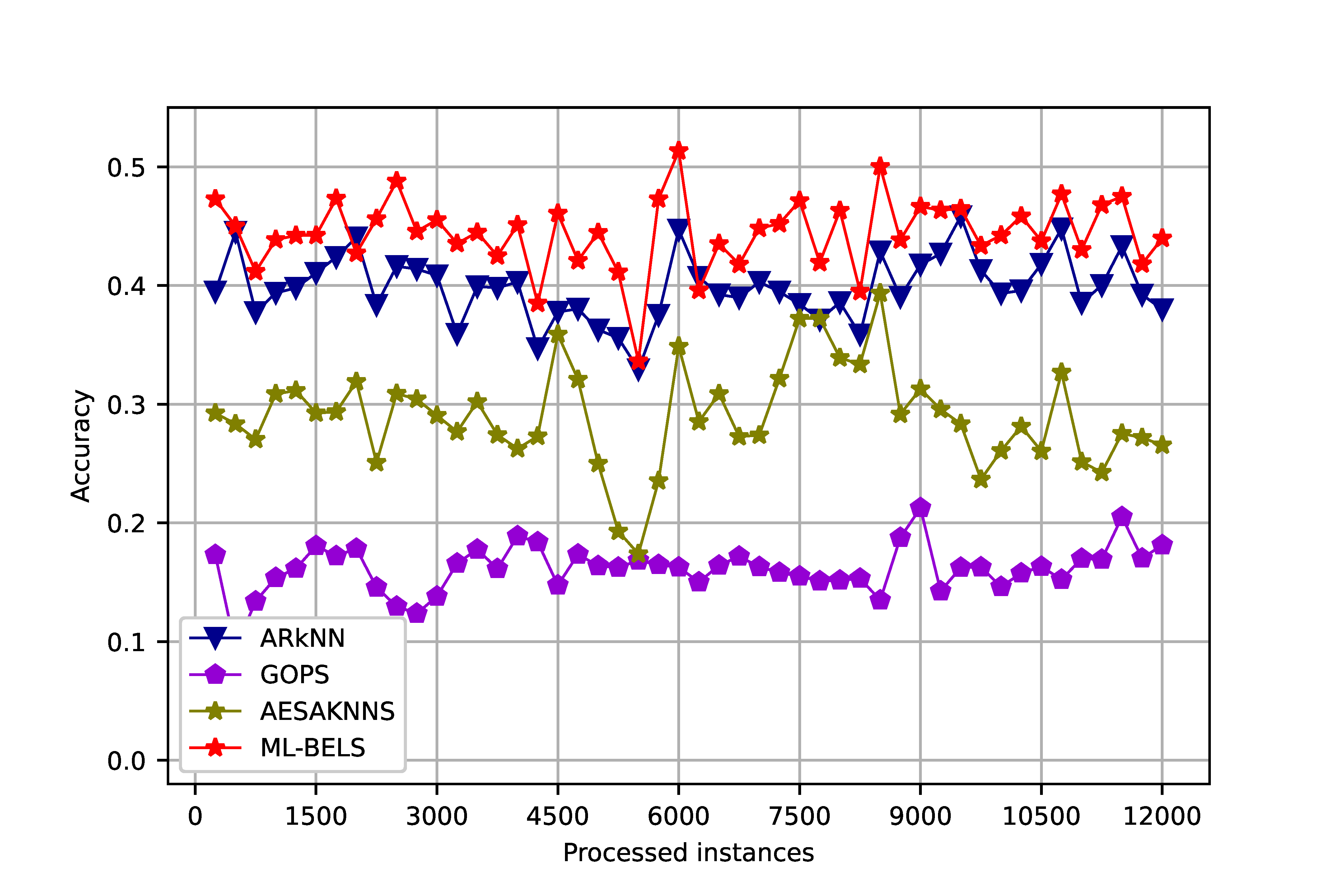} &
\includegraphics[width=0.3\columnwidth]{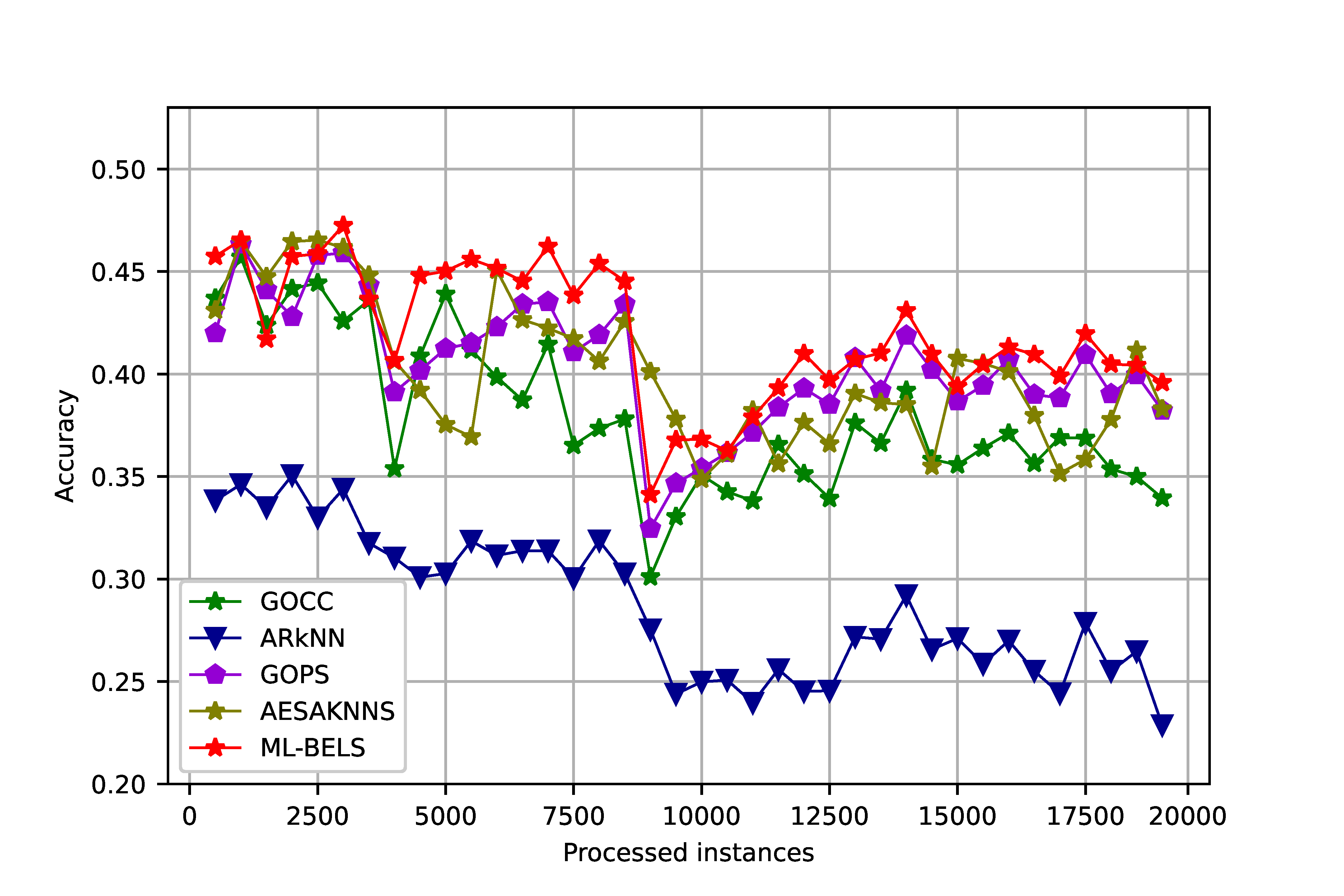} &
    
        \includegraphics[width=0.3\columnwidth]{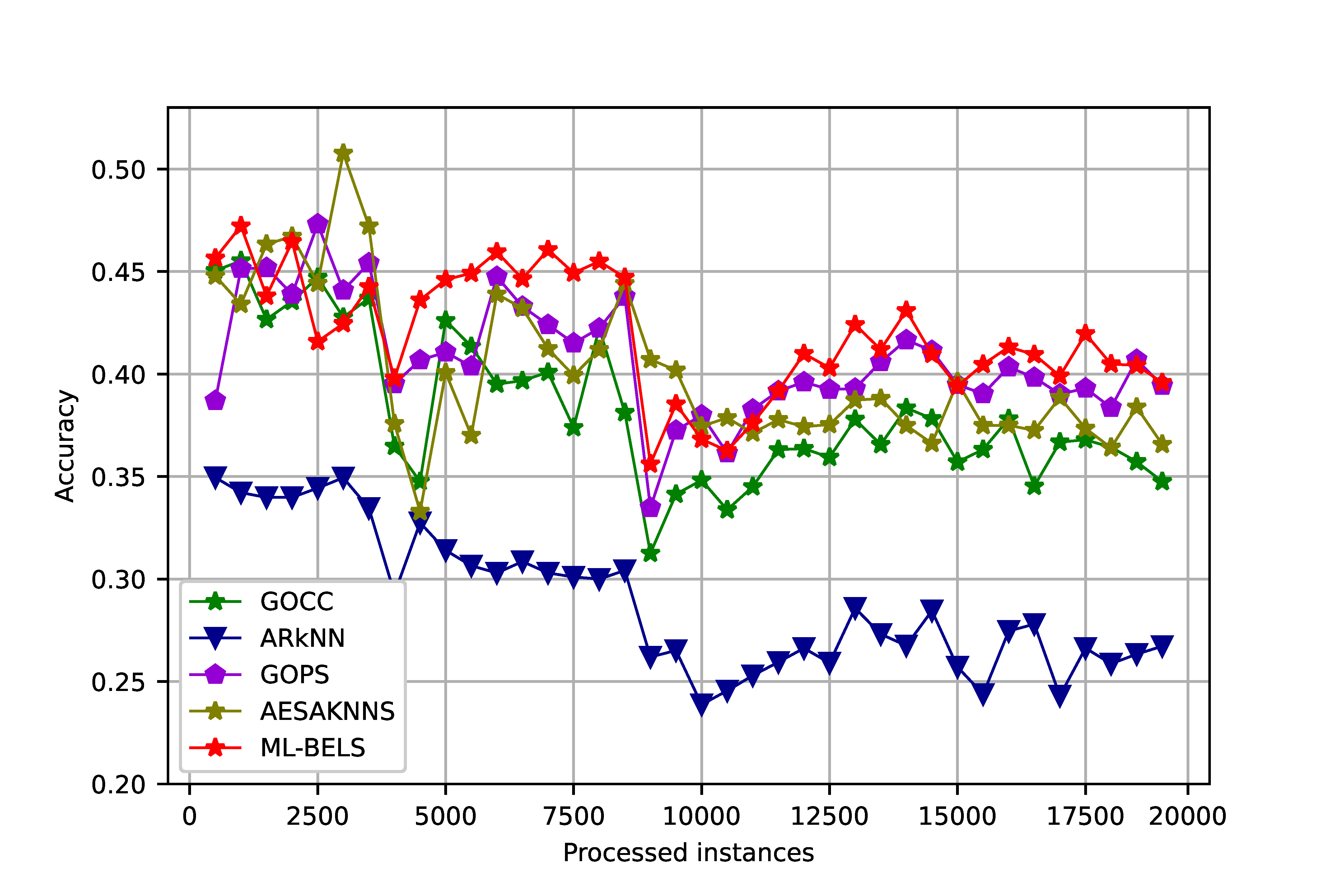} \\
    (d) Yahoo-Computers &
    (e) A-10-20&
    (f) G-10-10 
    \\

\end{tabular}
    \caption{Prequential temporal accuracy results of four real and two synthetic datasets. GOCC results for the Yahoo-Computers dataset are not available and thus are not included in the corresponding plot (d).}
    \label{fig:preq_fig}
\end{figure*}

\subsection{Effectiveness and Efficiency Analysis}

\par
The experimental evaluation results are presented in Table \ref{tbl:accuracy}. Compared to the baseline models, our proposed approach has a better average performance and rank in all three effectiveness measures: accuracy, F1-score, and Micro-average F1. Regarding average runtime, our model with 10\% labeled data achieves the best results.
The results in this table demonstrate that our model has a more stable runtime compared to the baselines, and changes in the feature and label set sizes have a low impact on the efficiency of the model.
In the same table, the results for GOCC, EBR, EABR, ECC, EaCC, and EBRT on Yahoo-Computers and Yahoo-Society datasets are not obtained due to memory overflow. This shows the inefficiency of these baseline models while dealing with high-dimensional data. Despite the baselines that deal with inefficiency in such a case, our model in the supervised setting is 31 and 40 times faster compared to the fastest baseline in Yahoo-Computers and Yahoo-Society datasets, respectively.

\par
The prequential accuracy plots for six datasets (four real and two synthetics) are given in Figure \ref{fig:preq_fig}. In these plots, an obvious gap between the accuracy results of the proposed approach and other baselines is observed in real datasets. In datasets with concept drift (\ref{fig:preq_fig}.e and \ref{fig:preq_fig}.f), despite a decline in the performance of our model in drift points, our model gets back to its previous state swiftly which results in better overall accuracy. The synthetic datasets are also noisy (check Section 5.1.2 for more details about the synthetic datasets), and the results on these datasets demonstrate the robustness of our model different from noise percentages. 
\par

\par

\begin{figure*}[h]
    \centering
    \begin{tabular}{c c}
    \includegraphics[width=0.45\columnwidth]{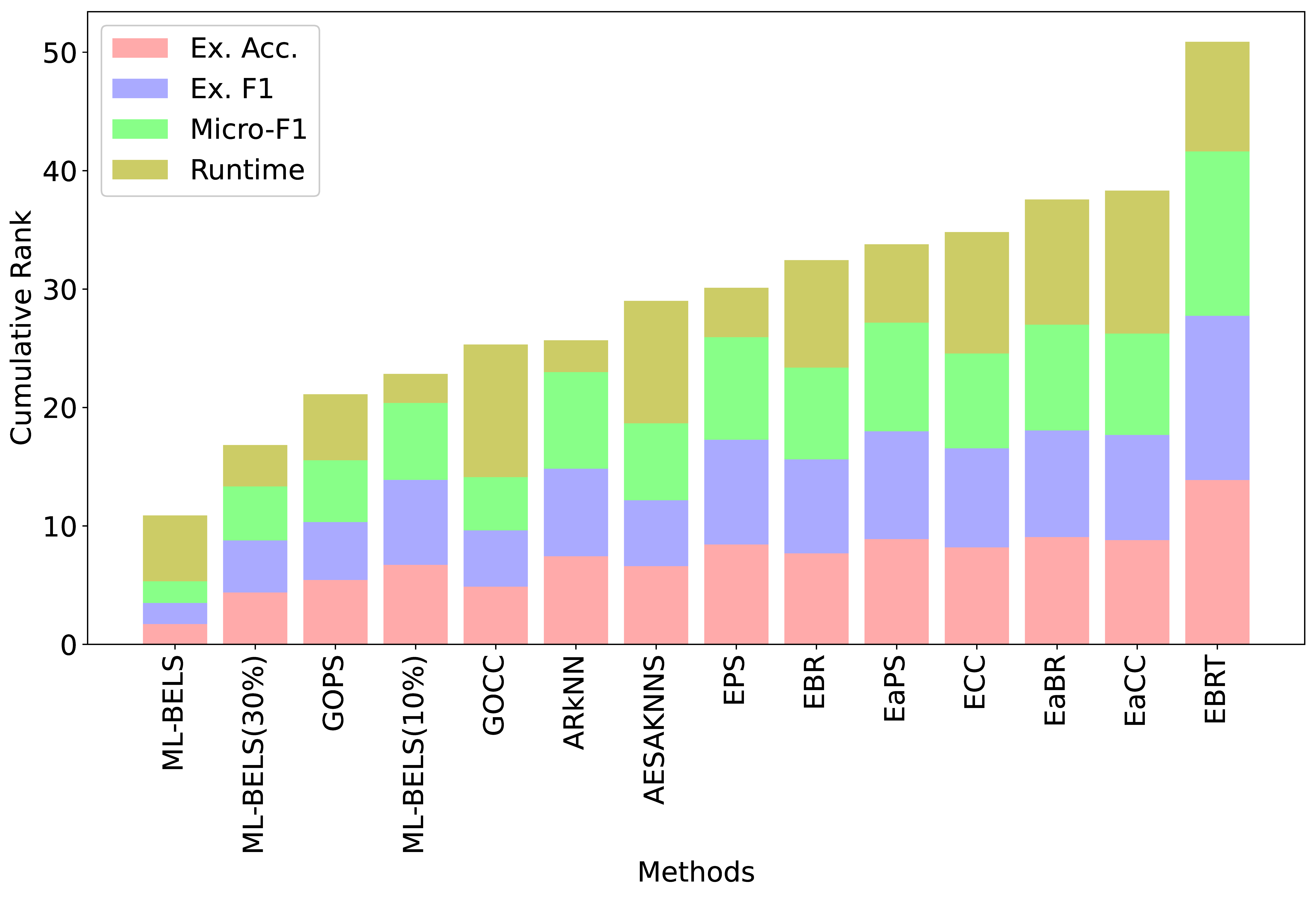}
    &\includegraphics[width=0.45\columnwidth]{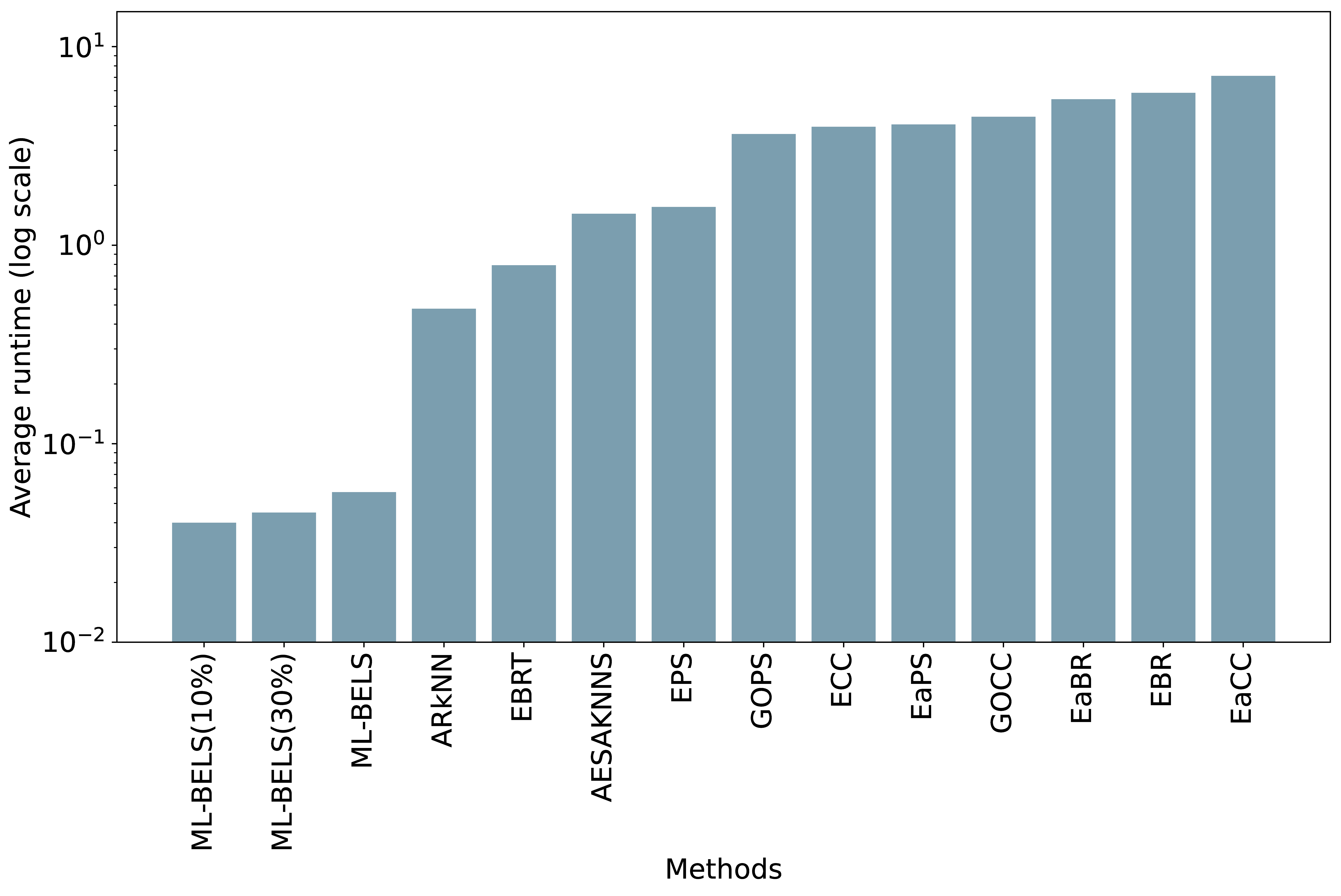}\\

    (a)&
    (b)

\end{tabular}
    \caption{(a) Cumulative rank of all metrics (lower is better) for the models. Models are sorted in terms of their cumulative rank. (b) Average runtime for processing 10 data instances in seconds (lower is better) in log scale. Models are sorted in ascending order in terms of their average runtime. Without taking the logarithms: Min (ML-BELS 10\%): 0.040, Max (EaCC): 7.120 seconds.}
    \label{fig:bars}
\end{figure*}

Regarding missing label handling, we report the results with two different portions of labeled instances and compare the results with the supervised models. In terms of average accuracy, F1-score, and Micro-F1, our model in the missing label setting beat the supervised approaches with only 30\% labeled data. Our model with 10\% labeled instances achieves the best average runtime and competitive results in the effectiveness metrics. The missing label settings reduce the computational burden of the model, and as expected, it has a lower runtime compared to the supervised ML-BELS. To provide a better understanding of the results and facilitate a more effective comparison, we present the cumulative ranks of the models (Figure \ref{fig:bars}.a), and the average runtime in log scale (Figure \ref{fig:bars}.b) in sorted order.

\par
Since our work is implemented in Python and the baselines are all in Java, comparing their memory usage would be unfair due to the differences in their memory management techniques; therefore, we report the memory usage of our model to show that it is efficient enough in terms of memory usage. Our model has a minimum usage of \textbf{2.397 MB} in the CHD49 dataset, and a maximum of \textbf{10,044.810 MB} in the Yahoo-Computers dataset. As we can observe, even when dealing with high-dimensional datasets, our model utilizes a reasonable amount of memory that is well-suited for most personal computers.

\par
\textbf{Limitations.} In binary relevance, a classifier is assigned for each label. An increase in the number of class labels results in more classifiers and thus a less efficient model. As for any approach based on binary relevance, our model is also affected by the large label set size. In Section 5.2, we demonstrate that the impact of larger label set sizes is minuscule; however, very large label sets may still adversely affect the efficiency of the model. In the missing label scenario, we halt the training of binary relevance components with missing labels, and as we can see in Table \ref{tbl:accuracy}, using 10\% percent of the data, our model is approximately 1.4 times (0.210 vs. 0.150) faster compared to its fully supervised version in terms of average runtime. Although putting a set of selected binary relevance components into a halt may alleviate the problem to a degree, it still cannot completely solve the issue in the case of a very large label set size.

\par

\begin{figure*}[h]
    \centering
    \begin{tabular}{c c c c}
    \includegraphics[width=0.23\columnwidth]{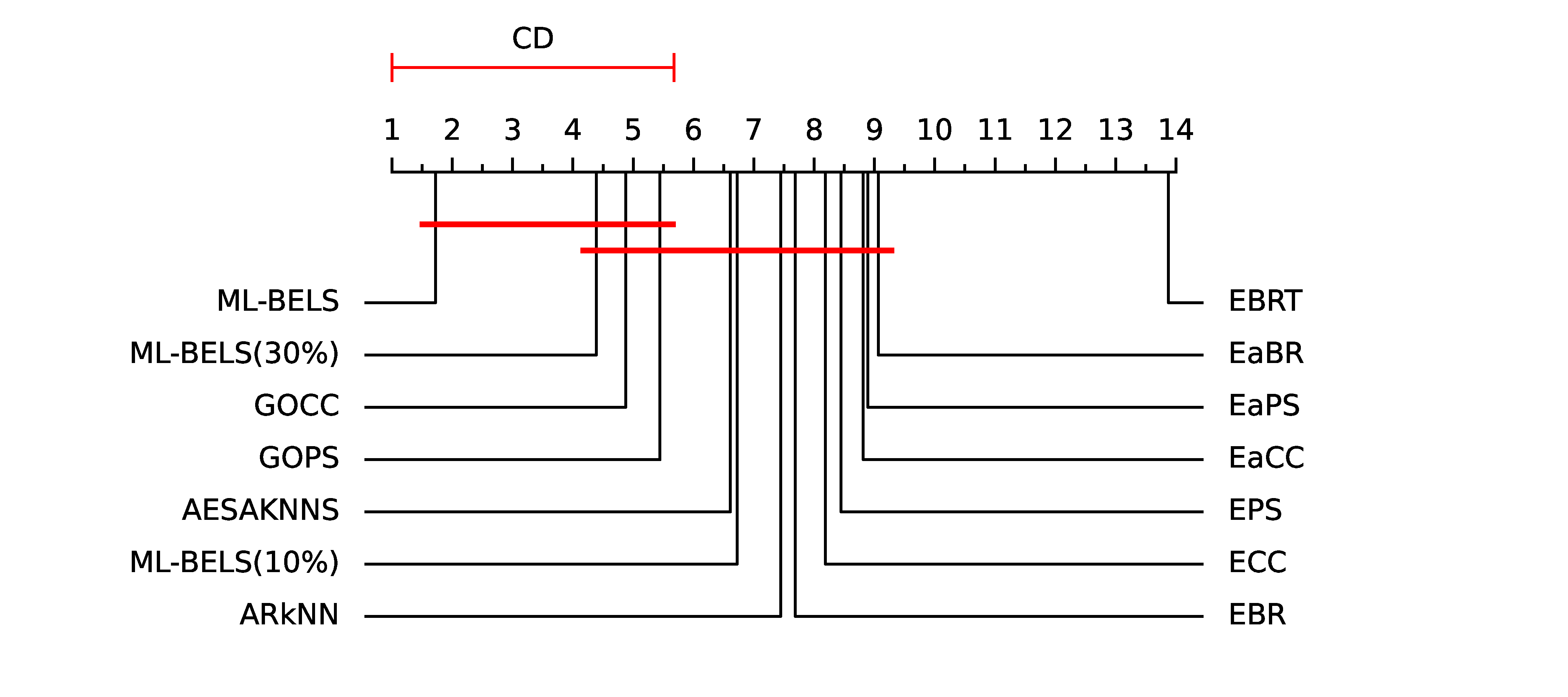}
    & 
    \includegraphics[width=0.23\columnwidth]{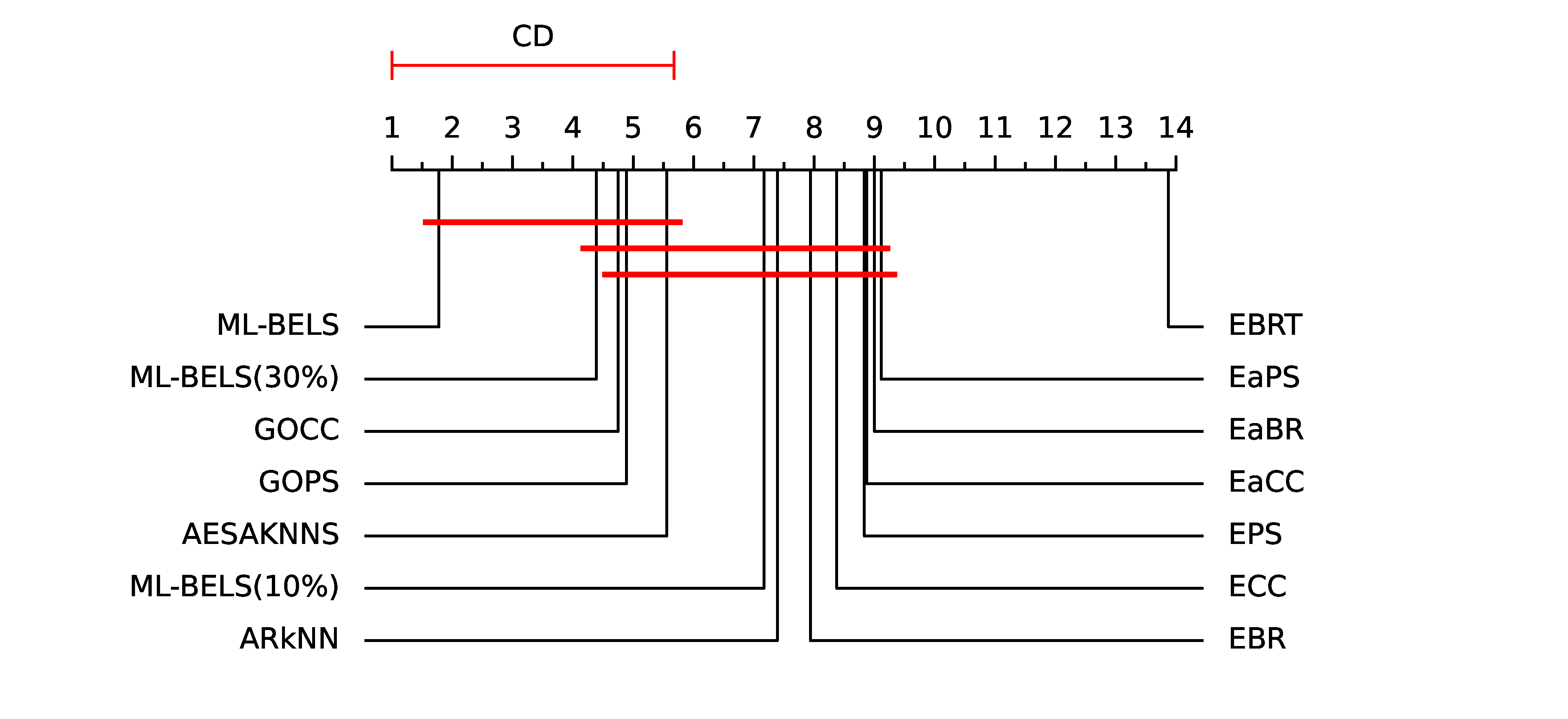}&
    
    \includegraphics[width=0.23\columnwidth]{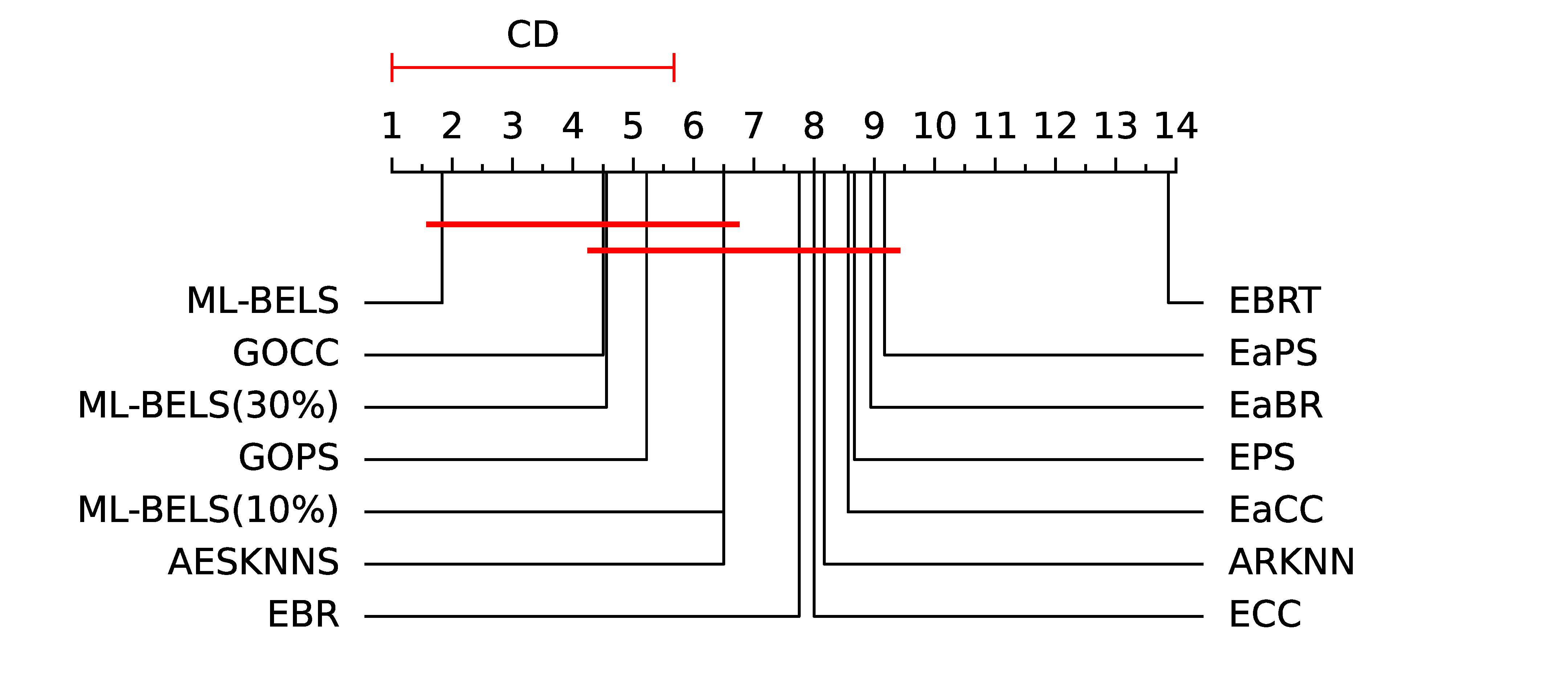} & \includegraphics[width=0.23\columnwidth]{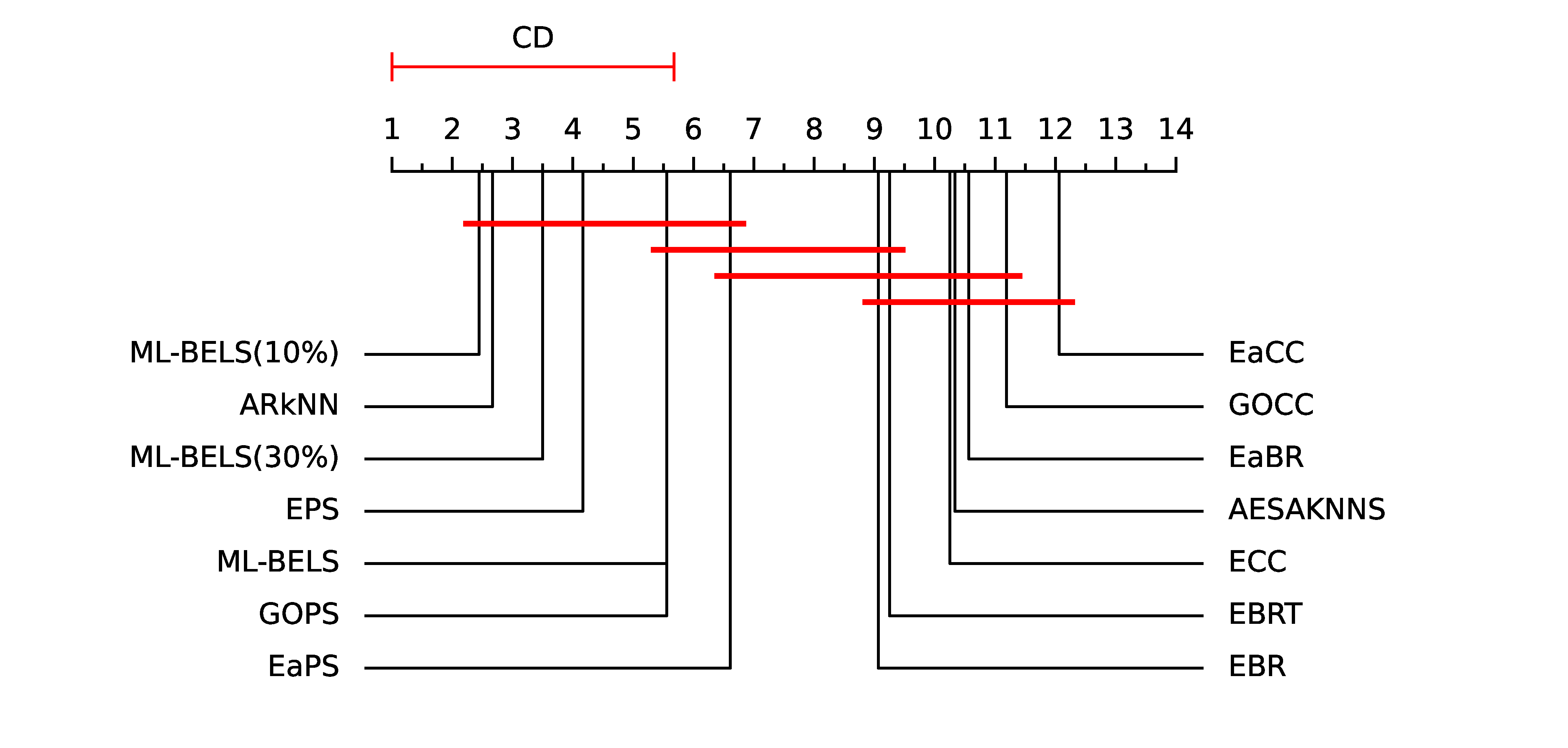} \\
    (a) Example-based accuracy &
    (b) Example-based F1-score&
    (c) Micro-F1  &
    (d) Runtime \\

\end{tabular}
    \caption{Critical distance (CD) diagram based on the ranks of the models (Tables \ref{tbl:accuracy}) for each evaluation metric (CD = 4.676). }
    \label{fig:stat}
\end{figure*}

\par
\subsection{Statistical Analysis}
Critical distance diagrams are provided in Figure \ref{fig:stat}. We apply the Friedman Test ($\alpha$= $0.05$) to reject the null hypothesis and then use the Nemenyi posthoc test (critical distance = 4.676) \cite{demvsar2006statistical}. The results in Figure \ref{fig:stat} show that our proposed approach (ML-BELS) achieves the best ranking in all four metrics of effectiveness and efficiency.
\par
In terms of example-based accuracy our model and its variant with 30\% missing labels are in the same group with GOCC, and GOPS, and statistically significantly outperform the rest of the baselines. In example-based F1-score, ML-BELS in supervised and missing label setting with 30\% missing data is in the same group with GOCC, GOPS, and AESAKNNS, and statistically significantly outperforms the remaining baselines. In Micro-F1, our model in both supervised and missing label settings is in the same group with GOCC, GOPS, and AESAKNNS, and statistically significantly outperforms the rest of the baselines. Regarding runtime, our model in both supervised and missing label settings is in the same group with ARkNN, EPS, GOPS, and EaPS, and statistically significantly outperforms the other baselines.

\subsection{Ablation Analysis}
In the ablation analysis, we study the effect of the weighting mechanism and ensemble. For brevity, two real and two synthetic datasets with a great variety of label cardinality and concept drift are chosen. We compare our model in default mode with three of its variants: ML-BELS with (1) only binary relevance (BR), (2) binary relevance and ensemble (BR+Ens), and (3) binary relevance, ensemble, and weighting altogether (BR+Ens+W).
\par
\begin{figure*}[t]
    \centering
    \begin{tabular}{c c c c}
    \includegraphics[width=0.23\columnwidth]{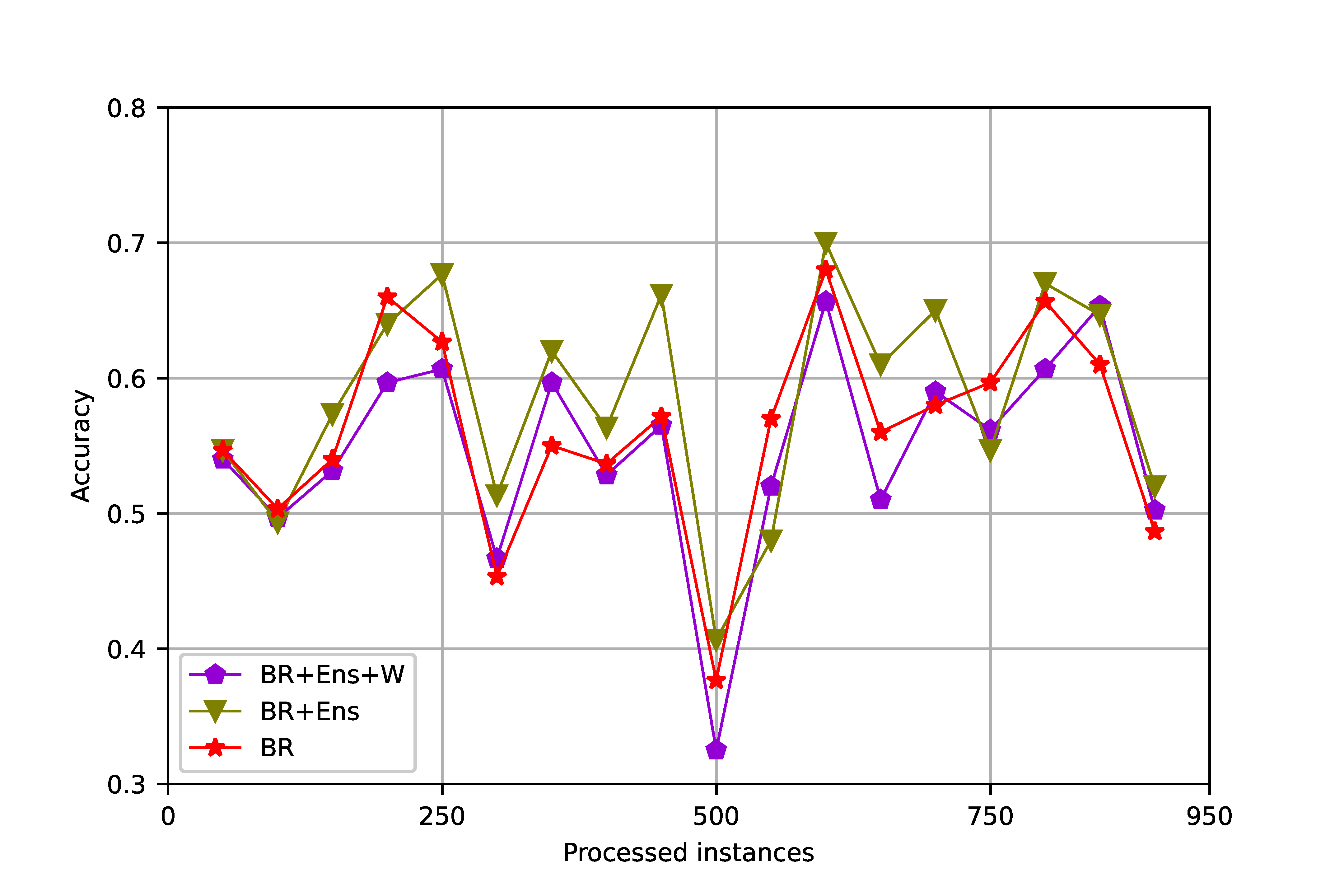}
    &\includegraphics[width=0.23\columnwidth]{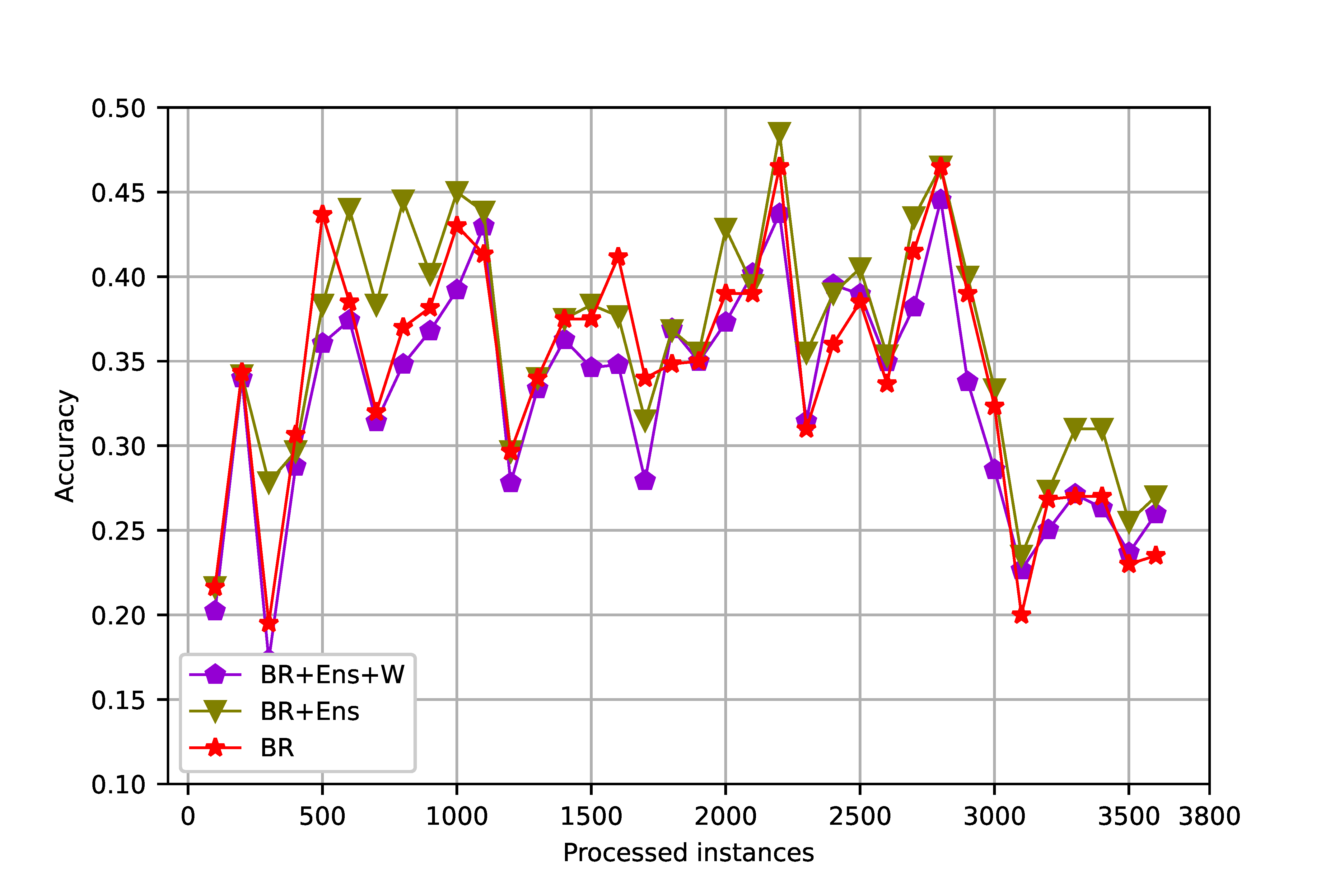}&

    \includegraphics[width=0.23\columnwidth]{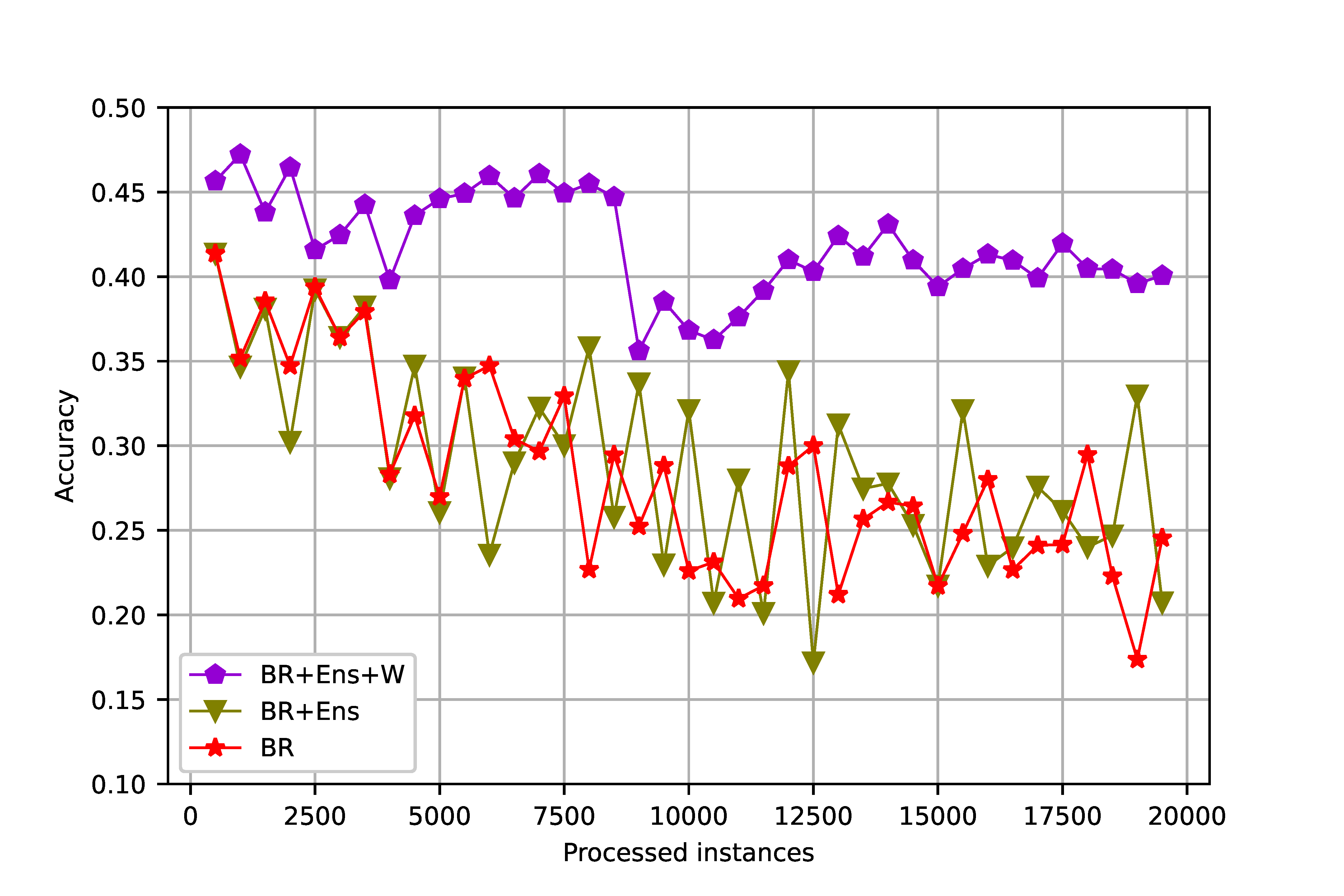}
    &\includegraphics[width=0.23\columnwidth]{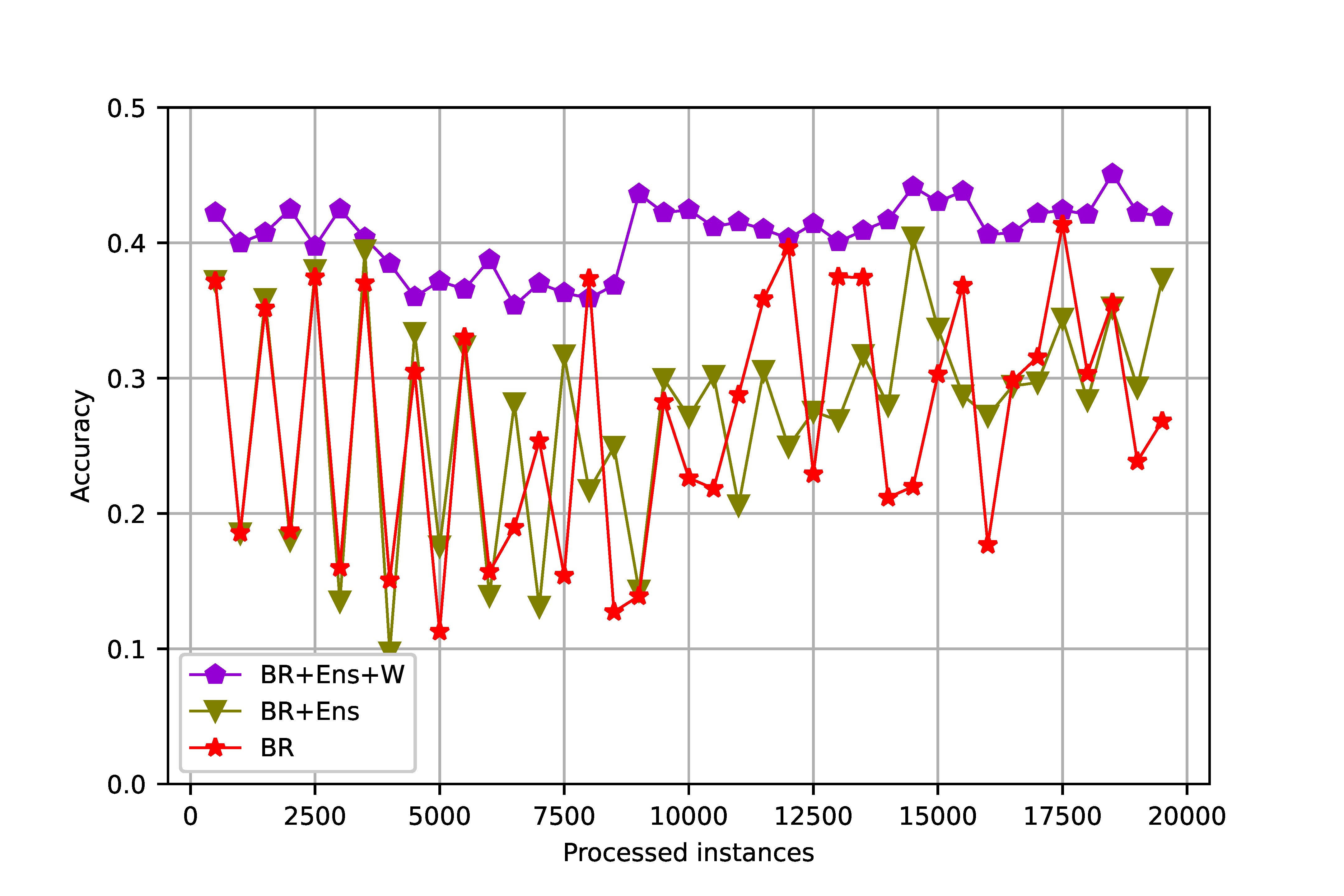}
    \\ 

    (a) Medical &
    (b) Slashdot&
    (c) G-10-10 &
    (d) R-A-15-10\\

\end{tabular}
    \caption{Prequential temporal accuracy results for ablation analysis. }
    \label{fig:preq_fig_ablation}
\end{figure*}

Table \ref{tbl:ablation} and Figure \ref{fig:preq_fig_ablation} report the ablation study results. BR mode achieves the lowest accuracy in all four datasets. Adding the ensemble approach improves the results, especially in datasets with low label cardinality. As mentioned earlier, we argue that our weighting mechanism has a negative impact on the accuracy of the model when applied to datasets with low cardinality and vice versa. The results support our claim and show that in datasets with high label cardinality, the weighting mechanism not only improves the results but also has a huge impact on handling the drift (Figure \ref{fig:preq_fig_ablation}.c and \ref{fig:preq_fig_ablation}.d). 

\begin{table}[t]
    \centering\footnotesize
    \caption{Ablation analysis example-based accuracy results}
    
    \begin{tabular}{l |r | r r r r }
    
        \hline 
         Dataset& Cardinality & BR & BR+Ens & BR+Ens+W  & Default \\
        \hline
        Slashdot  &1.181&0.342  &\textbf{0.361}&0.330 & \textbf{0.361}
      
        \\
        Medical &  1.245 & 0.561 &\textbf{0.584} & 0.547 & \textbf{0.584}
          \\
        G-10-10&4.352 &0.283 &0.291 &\textbf{0.419}&\textbf{0.419}\\
        
         A-R-15-10&3.378 &0.270 &0.275 &\textbf{0.405} &\textbf{0.405}\\

    \hline
    \end{tabular}
    
    \label{tbl:ablation}
\end{table}
\subsection{Hyperparameter Sensitivity Analysis}
We choose two main parameters and study their impact on two real and two synthetic datasets (Table \ref{tbl:para_sens}).
BELS achieves high accuracy by utilizing a large ensemble (default is 75). Since a large ensemble size slows down the data processing procedure, and our model consists of several ensembles (one for each label), a larger ensemble size increases the runtime. Based on this fact, we keep our ensemble size as small as three components for each ensemble. Besides, our results show that increasing the number of ensemble components does not contribute positively to the accuracy of the model. 

\par
The accuracy results for a threshold on LC are dependent on the label cardinality of the dataset. To instantiate, we have a high accuracy in two synthetic datasets with high cardinality until we reach the label cardinality of the dataset. In another example, we can see the effect of this threshold on two real datasets. With a low threshold ($\tau$= 1.2), the weighting mechanism is triggered and it results in lower accuracy. By increasing the threshold to more than the cardinality of the dataset, the accuracy increases.

\renewcommand{\arraystretch}{1}
\newcolumntype{M}[1]{>{\centering\arraybackslash}m{#1}}
\newcolumntype{P}[1]{>{\centering\arraybackslash}p{#1}}

\begin{table*}[h]
    \centering\footnotesize
    \caption{{Hyperparameter sensitivity analysis. Example-based accuracy and runtime in seconds for processing 10 data instances (in parenthesis) are provided. The best results for each hyperparameter are in bold. \textit{d} indicates the default value. }}

    \begin{tabular}{|l|M{10mm} M{10mm} M{10mm} M{10mm} M{10mm}|M{10mm} M{10mm} M{10mm}|}
    
    \hhline{~--------}
    \multicolumn{1}{l|}{}&
 \multicolumn{5}{c|}{ $\tau:$ LC threshold}&
    
    \multicolumn{3}{c|}{\textit{e}: Ensemble size}
    
    \\
        \hline
        Dataset & 1.2  & d= 1.5   & 2& 2.5& 4&d= 3  &10  & 20 \\
        \hline

        Slashdot&0.350 (0.030) & \textbf{0.361 (0.030)}&\textbf{0.361 (0.030)}&\textbf{0.361 (0.030)}&\textbf{0.361 (0.030)} &0.361 \textbf{(0.030)}& 0.362 (0.056)& \textbf{0.364} (0.082)\\

         Medical& 0.547 (0.075)&\textbf{0.584 (0.075})&\textbf{0.584 (0.075)}&\textbf{0.584 (0.075)}&\textbf{0.584 (0.075)}&\textbf{0.584 (0.075) }&0.574 (0.110) & 0.583 (0.120)\\
        
        A-R-15-10& \textbf{0.405 (0.016)}& 0\textbf{.405 (0.016)}& \textbf{0.405 (0.016)}&\textbf{ 0.405 (0.016)} & 0.362 (0.016)&\textbf{0.405 (0.016)} & 0.404 (0.029)& 0.403 (0.040)\\

          G-10-10& \textbf{0.419 (0.011)}&\textbf{ 0.419 (0.011)}& \textbf{0.419 (0.011)}& \textbf{0.419 (0.011) }& 0.291 (0.011)& \textbf{0.419 (0.011)}& 0.417 (0.019)& 0.418 (0.037) \\
         \hline
    \end{tabular}
    
    \label{tbl:para_sens}
\end{table*}

\section{Conclusion and Future Work}
In this paper, we propose a weighted binary-relevance-based model for multi-label classification in a data stream environment. Our results demonstrate that the proposed approach balances effectiveness and efficiency which results in a fast and accurate model regardless of the dimensionality of the data. We demonstrate that our proposed approach is robust to missing labels and concept drift.\par
One of the most important topics in multi-label classification is dealing with a large number of labels, a problem known as extreme multi-label classification. As future work, we plan to work on this problem in data stream classification and propose a solution based on ML-BELS for such settings.


\bibliographystyle{unsrt}  
\bibliography{references}

\end{document}